\documentclass[journal]{IEEEtran}

\IEEEoverridecommandlockouts

\usepackage[pdftex]{graphicx}
\usepackage{graphicx}
\usepackage{multicol}
\usepackage{multirow}
\usepackage{blindtext}
\usepackage{amsmath}
\usepackage{amssymb}
\usepackage{fancyhdr}
\usepackage{cite}
\usepackage[size=smaller]{caption}
\usepackage{array}                  
\usepackage{comment}

\usepackage{booktabs} 
\usepackage{tikz}
\usepackage[utf8]{inputenc}
\usepackage{pgfplots}
\usetikzlibrary{plotmarks}
\usepackage{pdflscape}
\pgfplotsset{compat=newest}
\pgfplotsset{plot coordinates/math parser=false}
\newlength\fwidth 

\usepackage{enumitem} 
\usepackage{algorithmic}
\usepackage[ruled,vlined]{algorithm2e}

\usepackage{textcomp}
\usepackage{subcaption}
\usepackage{hyperref}

\usepackage{xcolor}
\usepackage[export]{adjustbox}

\begin{document}

\title{AI-Enabled Covert Channel Detection in RF Receiver Architectures}

\author{\IEEEauthorblockN{Abdelrahman Emad Abdelazim, Alán Rodrigo Díaz-Rizo, Hassan Aboushady and Haralampos-G. Stratigopoulos\\} \IEEEauthorblockA{\small{Sorbonne Universit\'{e}, CNRS, LIP6, Paris, France}}\thanks{This work was supported by the Chips JU project Resilient Trust under Grant agreement N$^{\mbox{\scriptsize o}}$ 101112282.}}

\maketitle

\begin{abstract}

Covert channels (CCs) in wireless chips pose a serious security threat, as they enable the exfiltration of sensitive information from the chip to an external attacker. In this work, we propose an AI-based defense mechanism deployed at the RF receiver, where the model directly monitors raw I/Q samples to detect, in real time, the presence of a CC embedded within an otherwise nominal signal. We first compact a state-of-the-art convolutional neural network (CNN), achieving an 80\% reduction in parameters, which is an essential requirement for efficient edge deployment. When evaluated on the open-source hardware Trojan (HT)-based CC dataset, the compacted CNN attains an average accuracy of 90.28\% for CC detection and 86.50\% for identifying the underlying HT, with results averaged across SNR values above 1 dB. For practical communication scenarios where SNR $>$ 20 dB, the model achieves over 97\% accuracy for both tasks. These results correspond to a minimal performance degradation of less than 2\% compared to the baseline model. The compacted CNN is further benchmarked against alternative classifiers, demonstrating an excellent accuracy-model size trade-off. Finally, we design a lightweight CNN hardware accelerator and demonstrate it on an FPGA, achieving very low resource utilization and an efficiency of 107 GOPs/W. Being the first AI hardware accelerator proposed specifically for CC detection, we compare it against state-of-the-art AI accelerators for RF signal classification tasks such as modulation recognition, showing superior performance.

\end{abstract}


\section{Introduction}
\label{sec:intro}

\begin{figure}[t]
    \centering
    \includegraphics[width=1.0\linewidth]{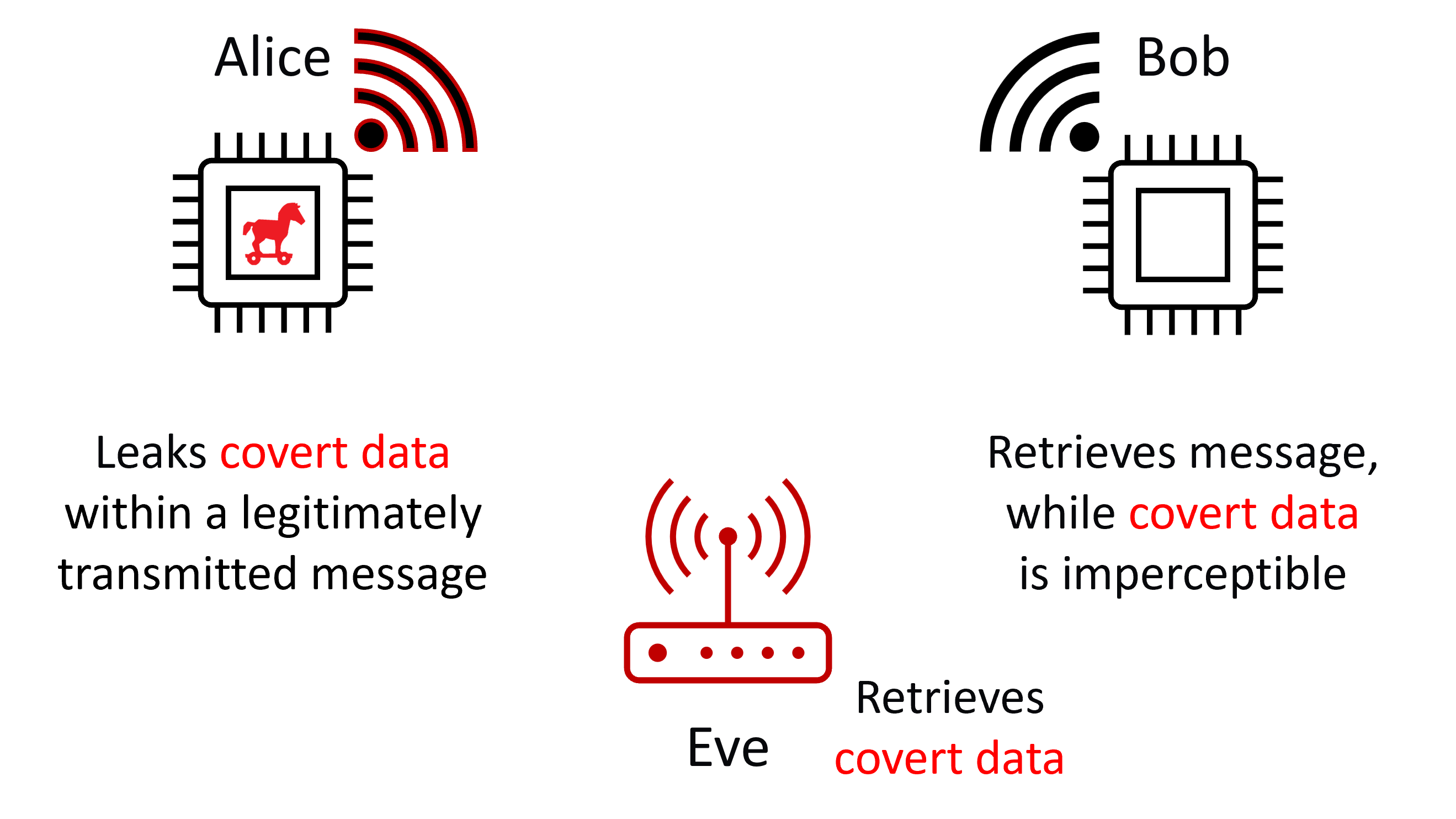}
    \caption{Covert channel threat model.}
    \label{fig:Threat_model}
\end{figure}

Covert Channels (CCs) represent a significant security threat for wireless chips. A CC refers to a stealthy communication channel that leaks sensitive information via the RF transmitter, such as encryption keys, cryptographic operations, configuration states, etc. Recent work has even demonstrated that entire AI models can be exfiltrated through CCs \cite{BD-RARA25}. In general, the attack works by routing the bits of the stolen information to the CC creation mechanism, which is responsible for embedding them into the RF transmission. Fig. \ref{fig:Threat_model} illustrates this threat model. The RF transmitter, a.k.a. Alice, is compromised with a CC creation mechanism that conceals sensitive information within an otherwise legitimate signal. The nominal RF receiver communicating with Alice, a.k.a. Bob, is inconspicuous and unaware of information leakage. A rogue RF receiver or eavesdropper, a.k.a. Eve, intercepts the transmission and, knowing how the CC operates, recovers the leaked bits. For instance, obtaining the cipher key would subsequently allow Eve to decrypt the communication. The leaked data is referred to as \textit{covert} data, whereas the legitimate message is referred to as \textit{cover} data.

The CC creation mechanism can itself be viewed as a Hardware Trojan (HT), which is why we use the term HT-based CC (HT-CC) in this work. HTs represent a broader threat model, encompassing any malicious modification of hardware \cite{TeKo10, KRRT10, BHBN14, XFJKBT16, BhTe18, HuWaChJiIEEEAccess20, JaZhGu21}. In general, a HT consists of two main components: a trigger and a payload. The trigger can be random, controlled, or always-on, while the payload, besides information leakage, may perform actions such as performance degradation or even denial of service. Beyond trigger and payload types, HTs can be classified by their insertion phase in the Integrated Circuit (IC) supply chain, e.g., during Electronic Design Automation (EDA) tools usage, design, integration of third-party IP blocks into a System-on-Chip (SoC), or fabrication, their insertion level, e.g., Register-Transfer-Level (RTL), gate-level, transistor-level, or layout, and their location on die, e.g., memory, processor, power management unit, analog and mixed-signal blocks, etc. For instance, in RF transceivers, a digital-to-analog attack \cite{EDiNSPLAS21,PPDiNFS22} is a plausible threat for inducing performance degradation or denial-of-service. In this scenario, a hidden trigger within a digital IP block of the SoC delivers a malicious digital word through the shared test infrastructure to the RF transceiver, thereby modifying its configuration. This type of attack can be detected at run time using lightweight on-chip monitors integrated into the RF transceiver \cite{PFLS22}.

Numerous works have showcased different implementations of HT-CCs for RF transceivers \cite{KKCR13,Dutta13,ClScHo15,HiFr10,GrSz13,SAANM19,DiRiAbSt22,GrPiKe21,JiMa10,LJNM17,SHANM20, CBOBO18,SRDJWA-SDMIC19,A-PN-PD-RLASCAS26}. Simultaneously, these works propose defense strategies to identify the HT-CC either at time zero using testing or on-line at run-time while the system is operating. A more detailed examination of existing HT-CC designs and their corresponding defenses is provided in Section \ref{sec:related_work}.

In particular, AI-based detection of CCs has proven highly effective \cite{D-RAAS24}. Since the covert data resides in the digital I/Q samples of each transmission frame, captured at the output of the RF receiver’s Analog-to-Digital Converters (ADCs), the approach is to encode these samples as ``images" and apply a Convolutional Neural Network (CNN) to identify the presence of a CC. An open-source hardware-generated HT-CC dataset, introduced in \cite{D-RAAS24} and detailed further in Section \ref{sec:dataset}, was used to train the CNN. This dataset contains transmission frames from RF transmitters with HT-CC capability and incorporates four prominent HT-CC techniques \cite{DiRiAbSt22,ClScHo15,Dutta13,SHANM20}, representing the major methods for embedding CCs into transmitted signals. Results showed that the CNN can not only distinguish CC-infected frames from CC-free ones but also identify the specific HT mechanism responsible for the creation of the CC.

Executing the CNN model directly on-chip is the ideal approach, as sending transmission frames to the cloud for inference raises privacy concerns, increases power consumption, and introduces latency in CC detection.

In this work, we advance beyond \cite{D-RAAS24} by designing a lightweight CNN model with $\approx 80\%$ fewer parameters and less than $2\%$ accuracy drop compared to the baseline CNN model in \cite{D-RAAS24}, enabling efficient on-chip execution. We also design a dedicated CNN hardware accelerator tailored to this task, suitable for integration into an RF transceiver for real-time CC detection under strict area and power constraints, and we demonstrate a first prototype on an FPGA.

The remainder of the article is organized as follows. Section \ref{sec:related_work} reviews prior work on HT-CC attacks and existing defense mechanisms. Section \ref{sec:dataset} introduces the HT-CC dataset, detailing the implemented attacks and data acquisition process. In Section \ref{sec:Defense_CC}, we motivate the choice of on-chip CNN execution and describe the RF receiver architecture featuring the integrated CNN hardware accelerator. Section \ref{sec:CNN} outlines all evaluated classifiers and elaborates on the design decisions behind the proposed compact CNN model. Section \ref{acceleratorarchitecture} presents the architecture of the CNN hardware accelerator. Section \ref{results} reports the HT-CC detection performance, and Section \ref{compare_with_others} compares our accelerator with state-of-the-art solutions. Finally, Section \ref{sec:conclusion} concludes the paper.

\section{Prior Art on HT-CCs}
\label{sec:related_work}

\begin{table*}[t]
\centering
  \caption{HT-CC attack models and defenses.}
  \begin{tabular}{p{1.38cm}p{6.9cm}p{6.9cm}p{1.19cm}}
    \toprule
    Ref. & Attack model & Defense & Present in the dataset\\
    \midrule
    \cite{KKCR13}
        & 
        Modifies the MAC layer Carrier Sense Multiple Access with Collision Avoidance (CSMA/CA) protocol to leak data into the timings of the transmitted packet sequence.
        & 
        Evades statistical tests that detect covert timing channels. No other defense is studied.
        & \\
    \cite{Dutta13,GrPiKe21} 
        & Encodes leaked data on the I/Q mapping and hides the encoding by introducing imperfections to the transmitted signal.
        & Certain tests, such as EVM, show a distinguishing behavior compared to HT-free operation; detectable by the AI-based defense \cite{D-RAAS24}.
        & \checkmark \\
    \cite{ClScHo15}-1 
        & Leaks data by introducing an additional phase shift into all STS symbols of the preamble.
        & Analysis of the preamble constellation; detectable by the AI-based defense \cite{D-RAAS24}.
        & \checkmark \\
    \cite{ClScHo15}-2 
        & Leaks data by introducing artificial CFO into each OFDM symbol. 
        & Analysis of CFO changes over time.
        & \\
    \cite{ClScHo15}-3, \cite{HiFr10} 
        & Leaks data in extra camouflage subcarriers added to the OFDM signal.
        & Decode the signal field to determine if the number of subcarriers is correct.
        & \\
    \cite{ClScHo15}-4, \cite{GrSz13}
        & Leaks data into replaced parts of the OFDM Cyclic Prefix (CP).
        & Compare the last 16 samples of an OFDM symbol with its CP; spectral analysis.
        & \\
    \cite{CBOBO18}
        & Leaks data using spread spectrum techniques.
        & Spectral analysis.
        & \\
    \cite{A-PN-PD-RLASCAS26}
        & Leaks data via a chaos-based spreading code and embeds the key below the noise floor.
        & No defenses are studied.
        & \\
    \cite{SAANM19}
        & Leaks data by substituting some legitimate data in the FEC block.
        & Channel noise profiling.
        & \\
    \cite{SRDJWA-SDMIC19}
         & Leaks data into controlled artificial RF impairments.
         & No defenses are studied.
         & \\
    \cite{DiRiAbSt22}
        & Leaks data through amplitude modulation (denoted by $\alpha$) of some subcarriers in the STS of the preamble.
        & Detectable by the AI-based defense \cite{D-RAAS24} but evades any other known defense for $\alpha<$15\%.
        & \checkmark \\
    \cite{JiMa10,LJNM17,SHANM20}
        & Leaks data by modulating amplitude and/or frequency of transmitted signal.
        & SSCF; ACE; use hardware dithering as a prevention mechanism \cite{KLAM18}; checking compliance of an invariant side-channel fingerprint \cite{LVHM15}; detectable by the AI-based defense \cite{D-RAAS24}.
        & \checkmark \\
  \bottomrule
\end{tabular}
\label{tab:prior_art}\vspace{-0.4cm}
\end{table*}

Table \ref{tab:prior_art} provides a concise summary of existing HT-CC attack models and corresponding defenses. These will be analyzed next in more detail.

\subsection{HT-CC attack models}

The adversary can hide the HT inside the digital section of the RF transceiver, e.g., at the Medium Access Control (MAC) \cite{KKCR13} or  baseband physical (PHY)  layers \cite{Dutta13, ClScHo15, CBOBO18, SAANM19, SRDJWA-SDMIC19, GrPiKe21, DiRiAbSt22, A-PN-PD-RLASCAS26}. Inside the PHY layer, the HT can act on the preamble generation \cite{ClScHo15, DiRiAbSt22}, payload generation \cite{ClScHo15,Dutta13,GrPiKe21,SRDJWA-SDMIC19}, Orthogonal Frequency-Division Multiplexing (OFDM) symbol generation \cite{HiFr10,ClScHo15}, or Forward Error Correction (FEC) block \cite{SAANM19}, or can leverage spread spectrum techniques \cite{CBOBO18, A-PN-PD-RLASCAS26}. Another category of HTs acts on the Analog Front-End (AFE) of the RF transceiver \cite{JiMa10, LJNM17, SHANM20}. A short description of the attack models is given in the second column of Table \ref{tab:prior_art}. The HT-CC attack models that are included in the HT-CC dataset used to train the CNN model are indicated with a check mark in the last column of Table \ref{tab:prior_art} and will be described in more detail in Section \ref{sec:dataset}.

It should be noted that there also exist HTs that induce physical side-channels to convey secret information in short range \cite{LKGPB09}. Herein, we consider only HTs that leak information in a CC in wireless technologies.

\subsection{HT-CC defenses}
\label{HT-CC defenses}

The aforementioned works also assess resilience against various defenses, and in most cases identify at least one effective defense, as indicated in the third column of Table \ref{tab:prior_art}. All defenses listed in Table \ref{tab:prior_art} operate post-silicon, either at test time or run time. We categorize them into two primary groups: non-AI and AI-based HT-CC detection methods focusing specifically on detecting the CC within the received RF signal. In addition, we discuss two further categories: HT-CC prevention methods that impede the feasibility of the HT-CC attack, and generic HT defenses that apply to any HT threat and either detect the underlying HT mechanism at pre-silicon or post-silicon or prevent HT insertion.

\subsubsection{Non-AI HT-CC detection defenses}

These defenses can be divided into two sub-categories: standard measurements at test time that verify compliance with the communication protocol, and specialized techniques at run-time designed to detect HT-related activity.

Standard measurements include evaluating Signal-to-Noise Ratio (SNR), Error Vector Magnitude (EVM), Bit Error Rate (BER), checking adherence to spectral mask specifications, and analyzing I/Q constellation diagrams.

Specialized techniques at run-time include Adaptive Channel Estimation (ACE) \cite{SHANM20}, channel noise profiling \cite{SAANM19}, and invariant side-channel fingerprinting \cite{LVHM15}. The ACE defense leverages the slow-fading characteristics of indoor communication channels to distinguish between channel impairments and HT activity. Channel noise profiling involves characterizing the distribution of noise present in the communication channel. The goal is to identify any unexpected systematic components that could potentially be attributed to HT activity. Invariant side-channel fingerprinting consists of creating an invariant fingerprint on the power supply and continually checking its conformity concurrently with the normal operation, where non-compliance implies potential HT activity.

\subsubsection{AI-based HT-CC detection}

The non-AI HT-CC detection defenses discussed above are generally too complex to be applied systematically at run time in nominal RF receivers, as doing so would increase design effort, complexity, and cost. Most of these approaches also operate offline. They require prior data collection and off-chip processing, leading to CC detection latency. Moreover, each method typically targets only specific HT-CC attack models, meaning that multiple defenses must be combined to cover all known HT-CC attack types and provide maximum security, which further increases overall defense cost.

An alternative direction is to employ Machine Learning (ML) or AI techniques. Statistical Side-
Channel Fingerprinting (SSCF) \cite{LJNM17,SHANM20} was the first ML-based approach demonstrated for HT-CC detection. It consists of training a one-class classifier in a feature space composed of side-channel fingerprints or parametric measurements from golden HT-free devices. The HT-infected devices have a feature vector that lies outside the classification boundary and, thereby, can be distinguished from HT-free devices. In \cite{LJNM17,SHANM20}, features are collected on the transmitted power, then Principal Component Analysis (PCA) is applied for feature dimensionality reduction. A one-class Support Vector Machine (SVM) is trained to learn the trusted boundary enclosing the HT-free population. The evaluation metrics are false positives (i.e., CC-free transmissions classified as CC-infected) and false negatives (i.e., CC-infected transmissions that evade detection).

However, while it has shown to be efficient for HT-CCs systematically distorting the transmission power \cite{JiMa10,LJNM17}, it has failed to screen out infected devices for more complex HT-CCs, such as the HT-CCs presented in \cite{DiRiAbSt22,SHANM20}. The reason is that RF receivers employ channel estimation algorithms that bundle together any malicious disturbances introduced by a HT with the inherent non-idealities of the wireless channel, i.e., the HT activity will be estimated along with the channel conditions and will be equalized or neutralized. In general, process variations within the transmitter hardware, noise, and channel effects make it extremely challenging for the SVM classifier in SSCF to distinguish in a space of parametric measurements the covert channel from the actual transmitted signal for all HT-CCs attack models.

In \cite{D-RAAS24}, it was proposed instead to train a more complex ML model, in particular a CNN, on raw transmission data rather than on derived parametric measurements. The results on the dataset described in Section \ref{sec:dataset} showed that the CNN could not only detect CC activity, but also identify the specific HT mechanism embedded in the RF transmitter, achieving over 99\% detection accuracy in the SNR range of interest. Building on this prior work, we propose a compact CNN model with $\approx 80\%$ less parameters with respect to the baseline CNN model in \cite{D-RAAS24} at the expense of a small accuracy loss of $\approx 2\%$, and we designed a lightweight CNN hardware accelerator tailored for this task, demonstrating its implementation on an FPGA.

\subsubsection{HT-CC prevention defenses}

Proactive measures can be taken during the design phase aiming at preventing the insertion of HT-CCs into RF transceiver ICs. Techniques that can be used include encrypting the PHY layer to prevent man-in-the-middle attacks such as eavesdropping \cite{CJJSKSD17} and design obfuscation techniques, such as RF transceiver functionality locking \cite{DRLAS22, DiRiAbSt23a, M-ZD-RAP-MV-GS25} and layout camouflaging \cite{LSLAS21}. These design obfuscation strategies aim at obscuring circuit functionality, thereby making it significantly more difficult for an attacker to introduce the HT mechanism.

Another strategy is to challenge the operational principles of HTs, aiming to neutralize their impact. An example here is the hardware dithering technique proposed in \cite{KLAM18}.

\subsubsection{Generic HT detection and prevention methods}

There are numerous generic pre-silicon and post-silicon HT detection countermeasures proposed in the literature \cite{TeKo10, KRRT10,BHBN14,XFJKBT16,BhTe18,HuWaChJiIEEEAccess20,JaZhGu21}. These countermeasures target primarily digital circuits and are potentially applicable for HTs residing in the PHY layer of the RF transceiver, although they have not yet been investigated in this context.

At pre-silicon, the RF transceiver owner can analyze the final IC prior to fabrication to detect HTs inserted by a rogue employee, an untrusted EDA tool provider, or a third-party IP vendor whose IP is integrated into the IC. Techniques include: verification and structural analysis of third-party IP blocks \cite{ZhTe11}; simulation of test patterns using logic testing tools to expose the HT \cite{HJASKvD19}; specific simulation benches to magnify the effect of the HT \cite{SKAHKK20}; Information Flow Tracking (IFT) methods that track the propagation of sensitive data and verify that they do not reach unauthorized sites in the design \cite{JGDBM17}; Proof-Carrying Hardware (PCH) methods to ensure that the hardware implementation is equivalent to its design specification, thus leaving little space for malicious logic insertion \cite{GDJFM15}; extracting circuit netlist features and using ML techniques to differentiate between normal nets and HT-infected nets \cite{HYT_IOLTS17,GCCBH_ACM23}.

However, the HT can be inserted at a later stage by a malicious foundry, thus a post-silicon HT detection method is more inclusive in terms of possible threat scenarios. Reverse engineering can be used to identify modifications in the IC \cite{SSFTHSF15}, but this approach is destructive, thus limited to testing a small number of IC samples. At test time, one approach is to take advantage of the parametric variations caused by HT circuitry or HT activity aiming at exposing the HT by its effect on parametric measurements, i.e., power, temperature, electromagnetic profile, etc. \cite{ABKRS07, NYWMB12, BaFoSr15, HZGJ17, SMTGFT20}. This is the approach employed at run-time in SSCF, combined with ML, in particular an SVM, to distinguish HT-free from HT-infected devices.

\section{HT-CC Dataset}
\label{sec:dataset}

\subsection{Implemented HT-CC attacks description}
\label{sec:attacks_description}

Here, we describe the four HT-CC types, indicated with check marks in the last column of Table \ref{tab:prior_art}, that make up the dataset.

\subsubsection{Cover signal}
\label{subsec:cover_signal}

\begin{figure}[t]
  \centering
  \includegraphics[width=1\linewidth]{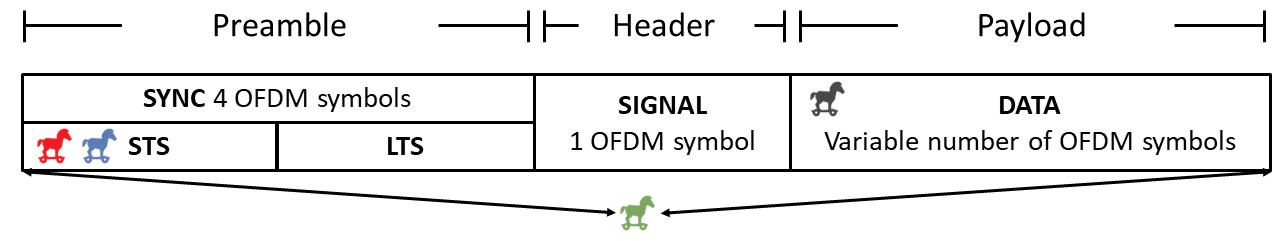}\vspace{-0.2cm}
  \caption{PPDU frame format of an OFDM IEEE 802.11, a.k.a. WiFi, transmission. The red, blue, black, and green Trojan horses depict the location of the covert message within the cover signal for HT1-CC, HT2-CC, HT3-CC, and HT4-CC, respectively.}
  \label{fig:PPDU_CC}\vspace{-0.0cm}
\end{figure}

The PHY cover signal consists of Physical layer Protocol Data Unit (PPDU) frames. The format of an IEEE 802.11 PPDU frame, as defined in the IEEE 802.11 standard (commonly known as Wi-Fi) \cite{ieee80211}, is shown in Fig. \ref{fig:PPDU_CC}. It is divided into three parts, namely preamble (a.k.a. SYNC), header (a.k.a. SIGNAL), and payload (a.k.a. DATA). The preamble section is composed of two different training sequences, namely a Short Training Sequence (STS) used for synchronization between the transmitter and the receiver, and a Long Training Sequence (LTS), used for estimating the center carrier frequency offset, as well as reference for generating the initial equalizer parameters. The header is composed of 1 OFDM symbol and describes the length of the packet, and the modulation and coding scheme of the following OFDM payload symbols. The payload is composed of several OFDM symbols depending of the type of transmitted frame. Each symbol is composed of subcarriers modulated either as BPSK, QPSK, 16-QAM, 64-QAM, or higher M-QAM constellations depending on the WiFi technology

\subsubsection{Covert message generation}
\label{subsec:covert_preamble}

\paragraph{Preamble modulation} 

\begin{figure}[t]
    \centering
    \begin{subfigure}{1\linewidth}
         \centering
         \includegraphics[width=0.6\textwidth]{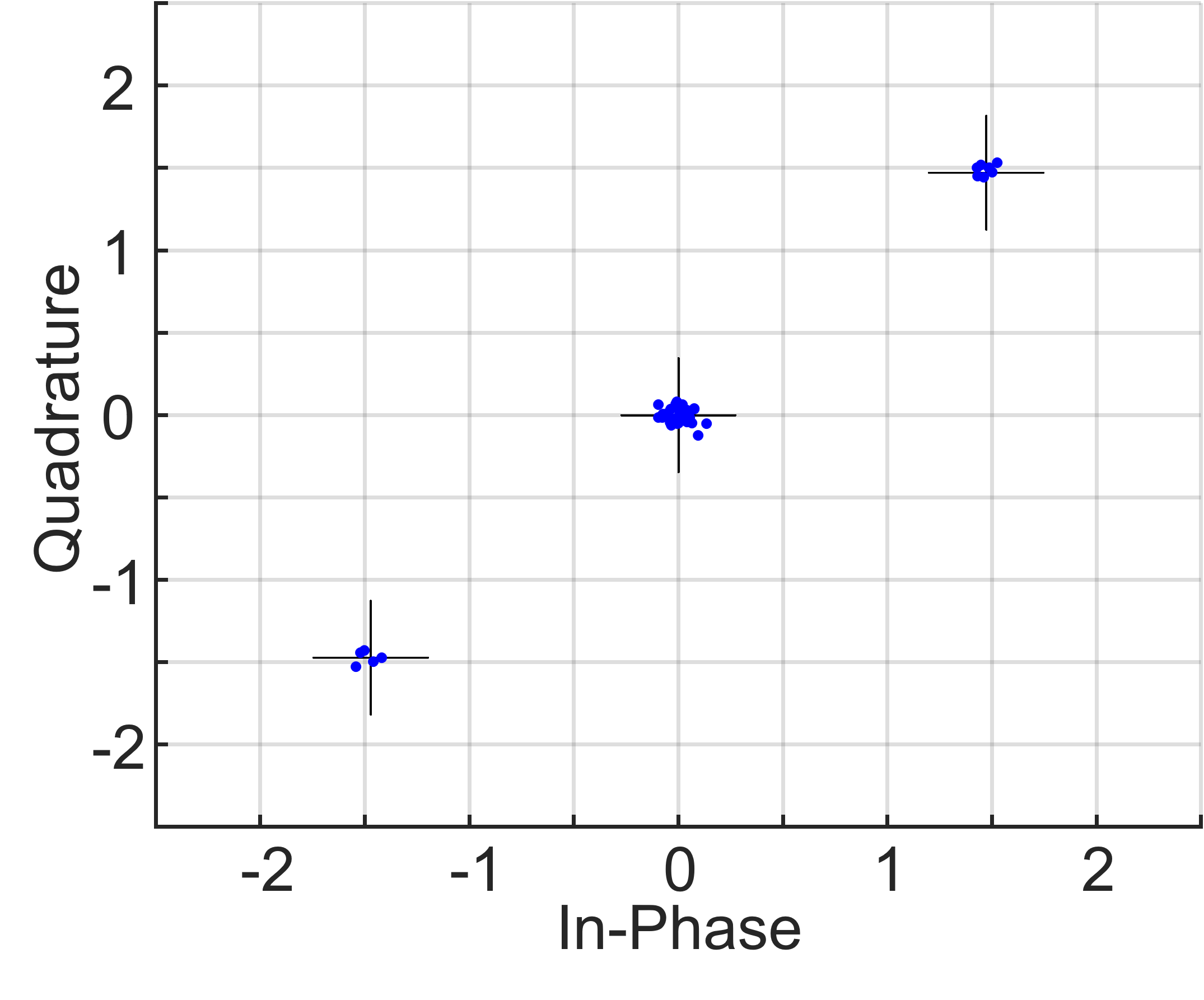}\vspace{-0.0cm}
         \caption{Nominal STS constellation diagram according to the standard \cite{ieee80211}.\label{fig:HT0}}
    \end{subfigure}\vspace{+0.2cm}
    \begin{subfigure}{1\linewidth}
        \centering
        \includegraphics[width=0.6\textwidth]{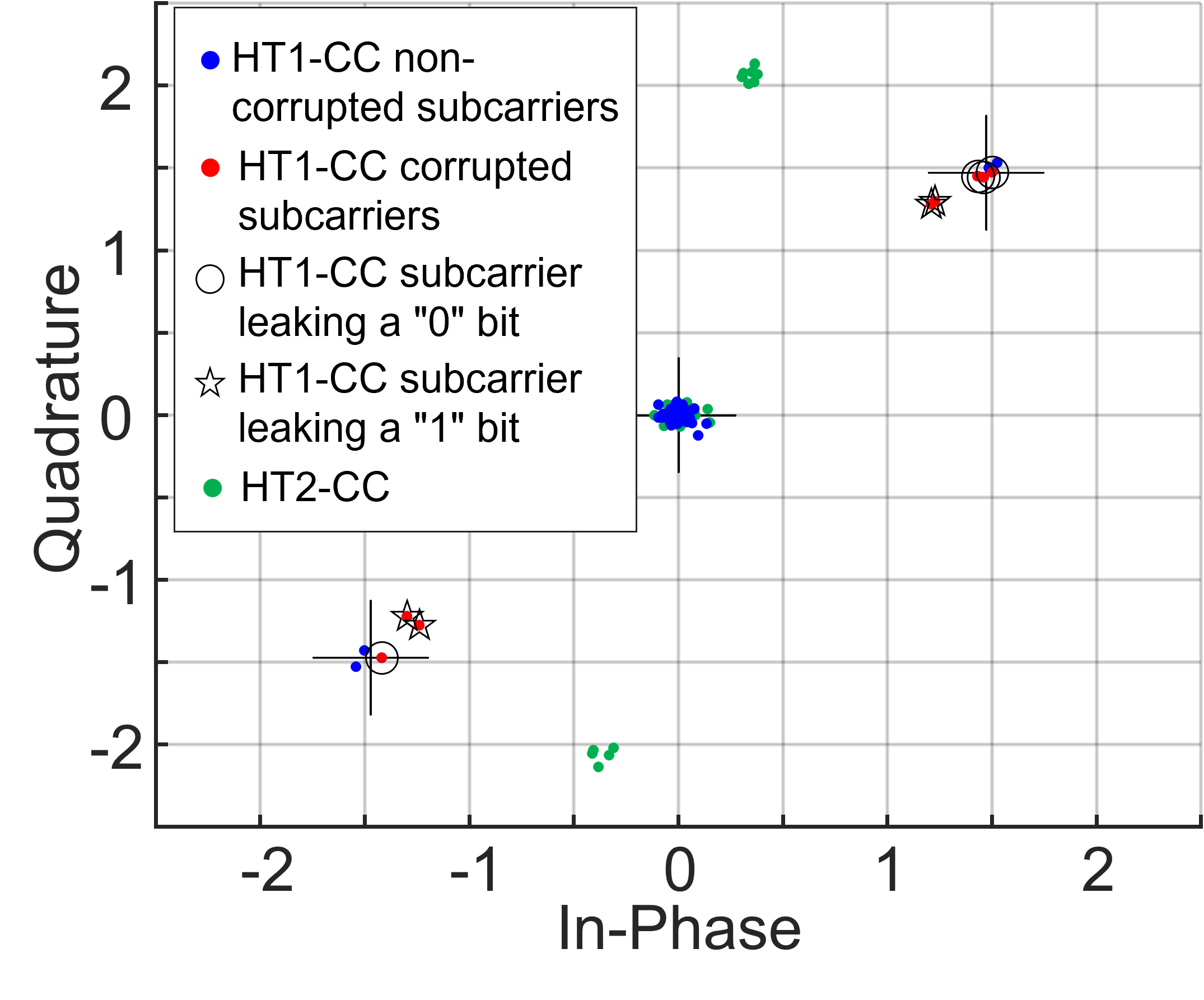}\vspace{-0.0cm}
        \caption{CC-infected STS constellation diagram in the case of HT1-CC \cite{DiRiAbSt22} and HT2-CC \cite{ClScHo15}. \label{fig:HT1_HT2}}
    \end{subfigure}\vspace{-0.0cm}
       \caption{Covert message hidden in the preamble. Additive White Gaussian Noise (AWGN) has been added to the signals such that the subcarriers are not superposed for illustration purposes.}
       \label{fig:CC_preamble} \vspace{-0.0cm}
\end{figure}

A first approach is to hide the HT inside the PHY layer and make it modulate the STS of the preamble of the transmitted frame, as shown with the Trojan horses placed in the STS field of the frame in Fig. \ref{fig:PPDU_CC}. The frequency-domain representation of the STS, denoted by STS\textsubscript{F}, is composed of 64 complex values, i.e., having real (I) and imaginary (Q) components, also called subcarriers or frequency bins, from the alphabet $\{-1.472-1.472j, 0, 1.472+1.472j\}$. The 64 subcarriers are indexed from -32 to 31, and there are 12 non-zero subcarriers, as shown in Table \ref{tab:STS}.
Fig. \ref{fig:HT0} shows a constellation diagram of the nominal STS\textsubscript{F} where the transmitted values, plotted as blue dots, fall near the expected constellation points, i.e., $\{-1.472-1.472j, 0, 1.472+1.472j\}$,  indicated by black plus signs. 
The time-domain representation of the STS, denoted by STS\textsubscript{t}, is derived by performing an Inverse Fast Fourier Transform (IFFT) on the STS\textsubscript{F}. The PPDU STS is composed of two OFDM symbols and it is obtained by concatenating two and a half STS\textsubscript{t}.
For simplicity, in the rest of the paper we will refer to STS\textsubscript{F} as STS.

\begin{table}[t]
\centering
  \caption{Frequency-domain representation of the STS (STS\textsubscript{F}) within the preamble of a WiFi frame.}
  \label{tab:STS}
  \begin{tabular}{p{3.9cm}p{3.9cm}}
    \toprule
    STS\textsubscript{F} index ($k$) & Complex-value (I,Q) \\
    \midrule
    -24, -16, -4, 12, 16, 20, 24 & 1.472 + 1.472j \\
    -20, -12, -8, 4, 8 & -1.472 - 1.472j \\
    others & 0 \\
  \bottomrule
\end{tabular}\vspace{-0.0cm}
\end{table}

The dataset considers two implementations of this attack type, referred to as HT1-CC and HT2-CC, shown with the red and blue Trojan horses in Fig. \ref{fig:PPDU_CC}, respectively. 

\begin{itemize}
    \item \textit{HT1-CC \cite{DiRiAbSt22}:} The data are leaked through minute amplitude modulations of non-zero subcarriers of the STS. These subcarriers are called \textit{corrupted subcarriers}. In particular, the amplitude of a corrupted subcarrier is reduced below a threshold level by multiplying it with $\alpha<1$ when the leaked bit is `1'. Otherwise, the amplitude is preserved for a leaked bit `0'. An amplitude modulation of 10\%, i.e., $\alpha=0.9$, is used and there are 8 corrupted subcarriers per STS, thus the CC leaks 8-bits per transmitted frame. Fig. \ref{fig:HT1_HT2} shows an example constellation diagram of the CC-infected STS where the leaked byte is composed of four `0' bits and four `1' bits.

    \item \textit{HT2-CC \cite{ClScHo15}:} The leaked data is encoded through a controlled counter-clockwise phase shift in all the STS subcarriers with respect to the CC-free constellation points. The number of possible phase shifts varies depending on the number of bits intended to be encoded and the shift amount in binary corresponds to the leaked byte. The CC implemented in the dataset leaks 8 covert bits per STS or per transmitted frame, resulting in 256 possible phase shifts. Fig. \ref{fig:HT1_HT2} shows an example constellation diagram.

\end{itemize}

\begin{figure}[t]
    \centering
    \begin{subfigure}{1\linewidth}
         \centering
    \includegraphics[width=0.68\textwidth]{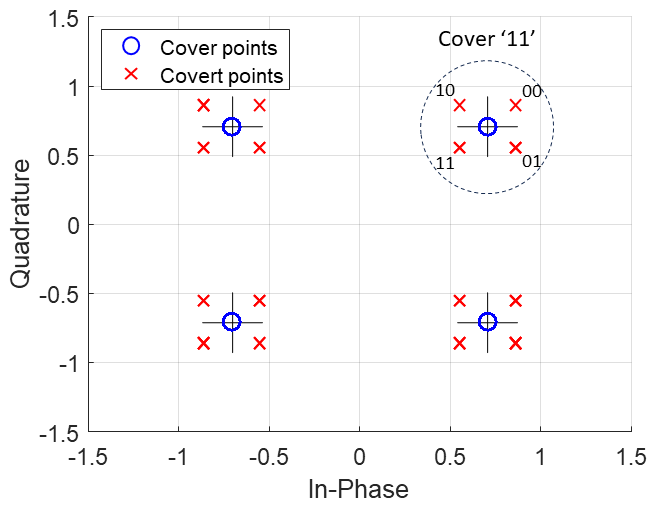}\vspace{-0.0cm}
         \caption{Possible initial displacement of the cover point to create the covert point.\label{fig:HT3a}}
    \end{subfigure}
    \begin{subfigure}{1\linewidth}
        \centering
    \includegraphics[width=0.68\textwidth]{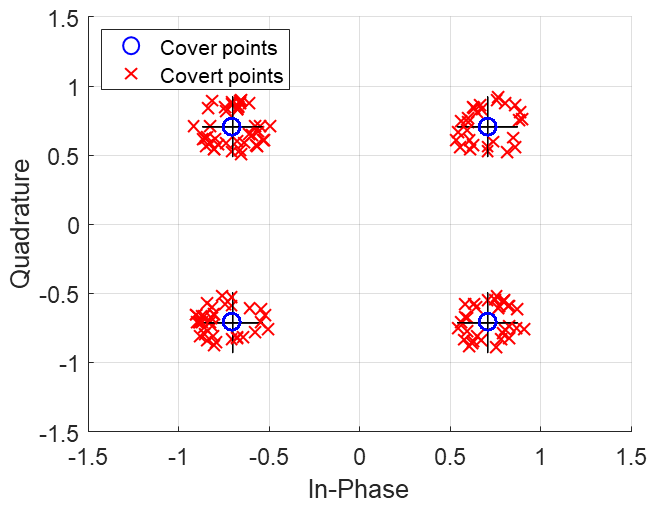}\vspace{-0.0cm}
        \caption{Possible final positions of the covert point.\label{fig:HT3b}}
    \end{subfigure}\vspace{-0.0cm}
       \caption{``Dirty" constellations.}
       \label{fig:HT3} \vspace{-0.0cm}
\end{figure}
\paragraph{Payload modulation}

\begin{itemize}
    \item \textit{HT3-CC \cite{Dutta13}:} A second approach, called ``dirty" constellations, is to hide the HT inside the PHY layer and make it modulate the covert message within the PPDU DATA field of the cover signal, as shown in Fig. \ref{fig:PPDU_CC} by a black Trojan horse. This approach takes advantage of hardware impairments and noisy channel conditions. More specifically, the PPDU DATA is composed of a variable number of OFDM symbols depending on the frame type and payload length. In the dataset implementation, the OFDM symbols comprise 48 QPSK modulated subcarriers. Let us consider the QPSK constellation diagram in Fig. \ref{fig:HT3a}. The attack can leak two bits per QPSK subcarrier. It chooses a subcarrier and displaces its constellation point according to the two bits being leaked. Essentially, using a QPSK cover point as origin there are four QPSK covert points. For example, if we want to leak bits `01' we can displace the upper right point (i.e., cover point `11') to the bottom right position (i.e., covert point `01'), as shown in Fig. \ref{fig:HT3a}. Then, several operations are applied to the covert point to reduce the probability of detecting the CC: (a) the covert point approaches symmetrically around the origin up to a distance equal to that of the 64-QAM; (b) its position is randomized with a Gaussian distribution within a dispersion radius $r=\sqrt{2/42}$; (c) it is rotated within $r$ with a monotonically increasing angle $\theta$ with steps of $15^o$. Moreover, to avoid detecting HT activity, even when the HT is inactive, intentional distortions are added to the transmitted signal increasing the average EVM up to 10 dB within the allowed limits of the IEEE 802.11 standard. Therefore, it is worth noticing that a device infected with HT3-CC will always have a higher BER compared to a CC-free device. Fig. \ref{fig:HT3b} depicts the outcome after these operations. In the dataset implementation, 5 ``dirty" subcarriers per OFDM symbol are used, thus 10 bits are leaked per OFDM symbol.

\end{itemize}

\begin{figure}[t]
  \centering
  \includegraphics[width=0.85\linewidth]{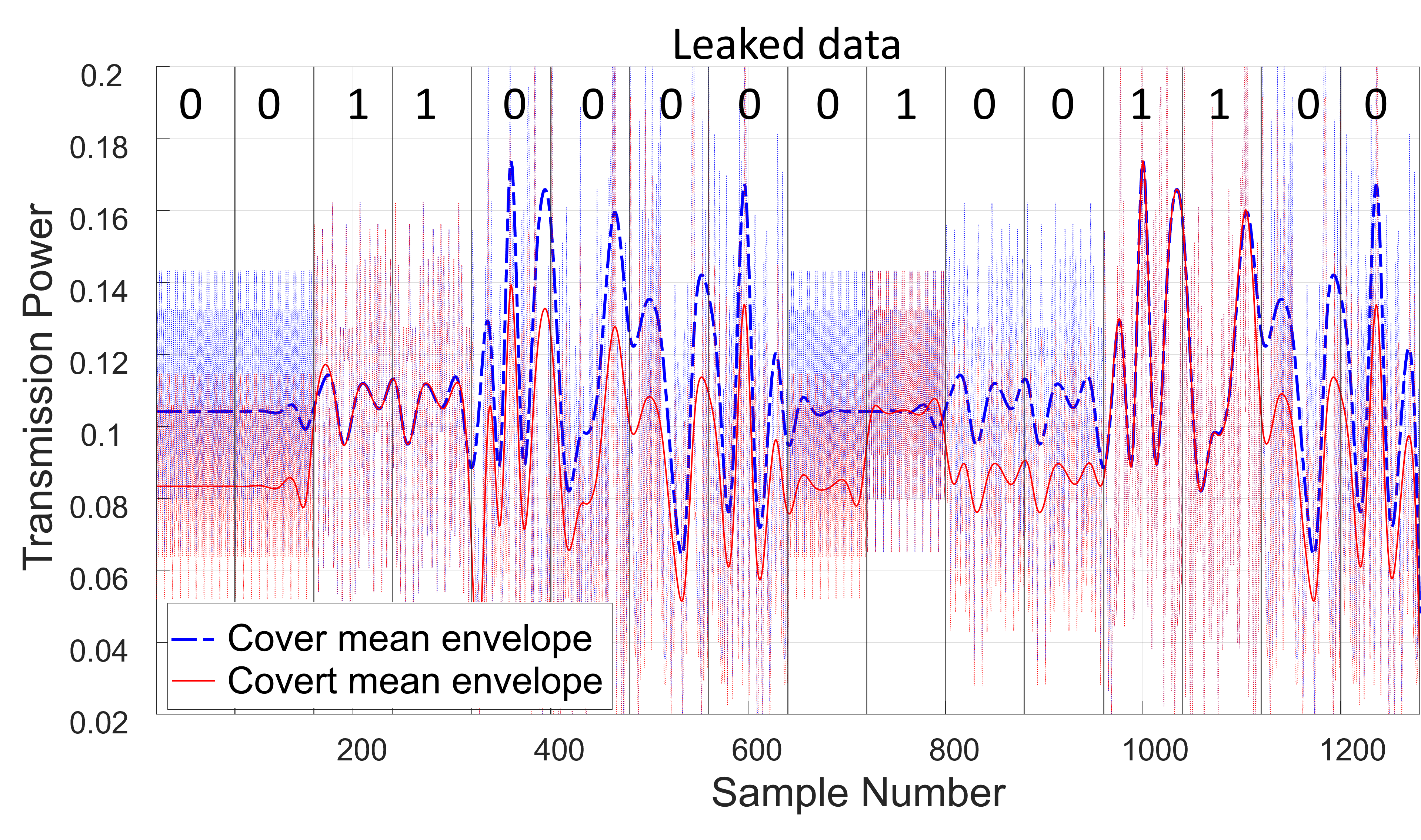}\vspace{-0.0cm}
  \caption{Mean envelope of the CC-free signal and the CC-infected signal with the HT4-CC attack. The signal mean envelope is reduced for a leaked bit `0', while it remains unchanged for a leaked bit `1'.}
  \label{fig:HT4} \vspace{-0.0cm}
\end{figure}
\paragraph{Power modulation}

\begin{itemize}
    \item \textit{HT4-CC \cite{SHANM20}:} The fourth considered attack model inserts the HT into the AFE leaking data through minute modifications in the amplitude of the transmitted signal. In this way, the covert message is spread across the transmitted frame, as shown with the green color Trojan horse in Fig. \ref{fig:PPDU_CC}. More specifically, RF transmitters use multiple programmable Variable Gain Amplifiers (VGAs) in the transmission chain to satisfy linearity and achieve desired performance specifications. A Serial Peripheral Interface (SPI) controls these VGAs through the PHY layer. The attack leaks secret information by systematically modifying the gain of the VGAs, creating minute variations in the transmit power, according to the leaked bits. In the dataset implementation, 8 bits are leaked per transmitted frame, where the amplitude is reduced by 20\% for a leaked bit `0', while it remains unchanged for a leaked bit `1', as illustrated in Fig. \ref{fig:HT4}. In \cite{SHANM20}, CC data rates of 2.5 and 5 bits per second (bps) are reported. However, these rates are significantly slower with respect to other HT-CCs in the dataset, making it challenging to perform fair comparisons. For instance, the HT1-CC exhibits a leakage of 8 bits per frame, resulting in a CC data rate of 28 kbps when considering a nominal number of frames per second (fps) of 3,500 at a data rate of 54 Mbps. This CC data rate is 5,600 times larger than that of the HT4-CC. Hence, it was decided to increase the throughput of the HT4-CC using a digitally-emulated version of the HT4-CC, leaking 8 bits per transmitted frame.
\end{itemize}

\subsection{HT-CC Dataset Acquisition}
\label{sec:dataset_acquisition}
\begin{figure}[t]
  \centering
  \includegraphics[width=1\linewidth]{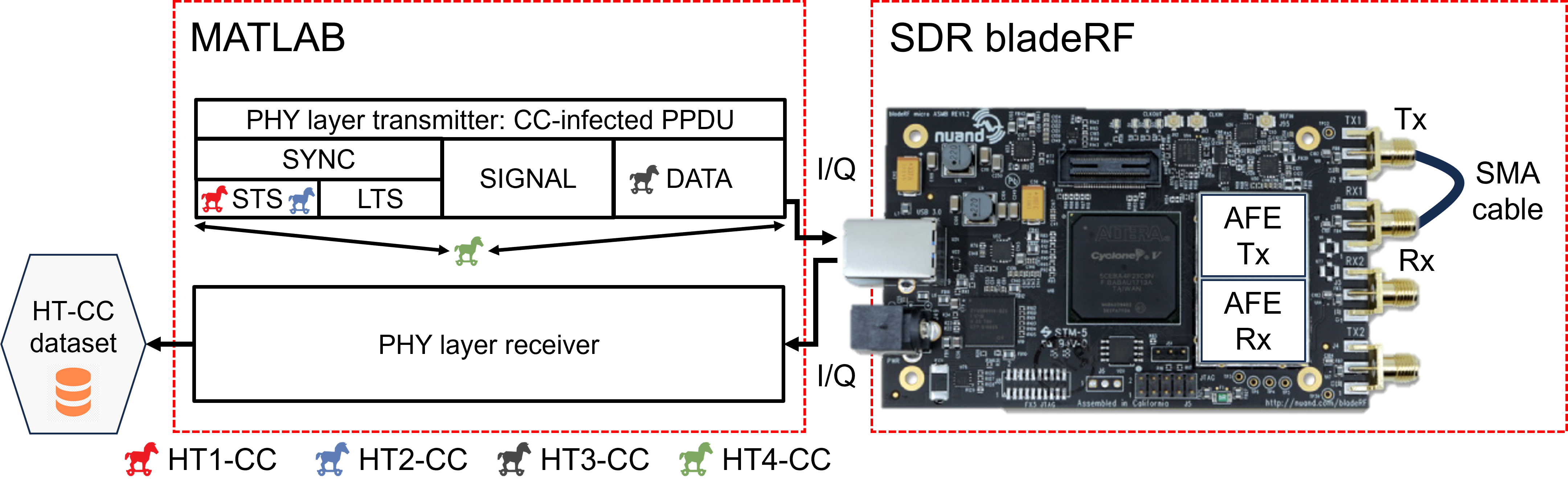}
  \caption{Experimental setup for dataset generation.}\vspace{-0.0cm}
  \label{fig:measurement_setup}
\end{figure}

The dataset is generated using the Software Defined Radio (SDR) bladeRF board from Nuand \cite{bladeRF}. The hardware acquisitions are performed using a single board, by connecting the RF transmitter with the RF receiver in loopback mode via an SMA cable, as shown in Fig. \ref{fig:measurement_setup}. The PHY layer implemented in MATLAB prepares the PPDU frames of the transmitted signal, shown in Fig. \ref{fig:PPDU_CC}, then the frames are transmitted using the RF transmitter of the board. While a CC-free transmission is composed of PPDU frames as defined by the IEEE 802.11 standard \cite{ieee80211}, a CC-infected transmission has a CC embedded into the PPDU frames leaking secret information as described in Section \ref{sec:attacks_description}.

The dataset is organized into 5 parts corresponding to the 4 HT-CCs described in Section \ref{sec:attacks_description}, denoted by HTX-CC, X=$\{1,\cdots,4\}$, and the CC-free signal. Each part consists of 16 elements representing 2 acquisitions with 8 different SNR values ranging from 1 to 29 dB with a step of 4 dB. Each element consists of 2000 fixed-length OFDM IEEE 802.11 frames, where each frame contains $2 \times 640$ real-value samples corresponding to the two I/Q branches.

The leaked message is common to all attack models. It is formed of random binary data, it has a length of 11 bytes, and it is repeated continuously.
\begin{figure}[t]
  \centering
  \includegraphics[width=0.85\linewidth]{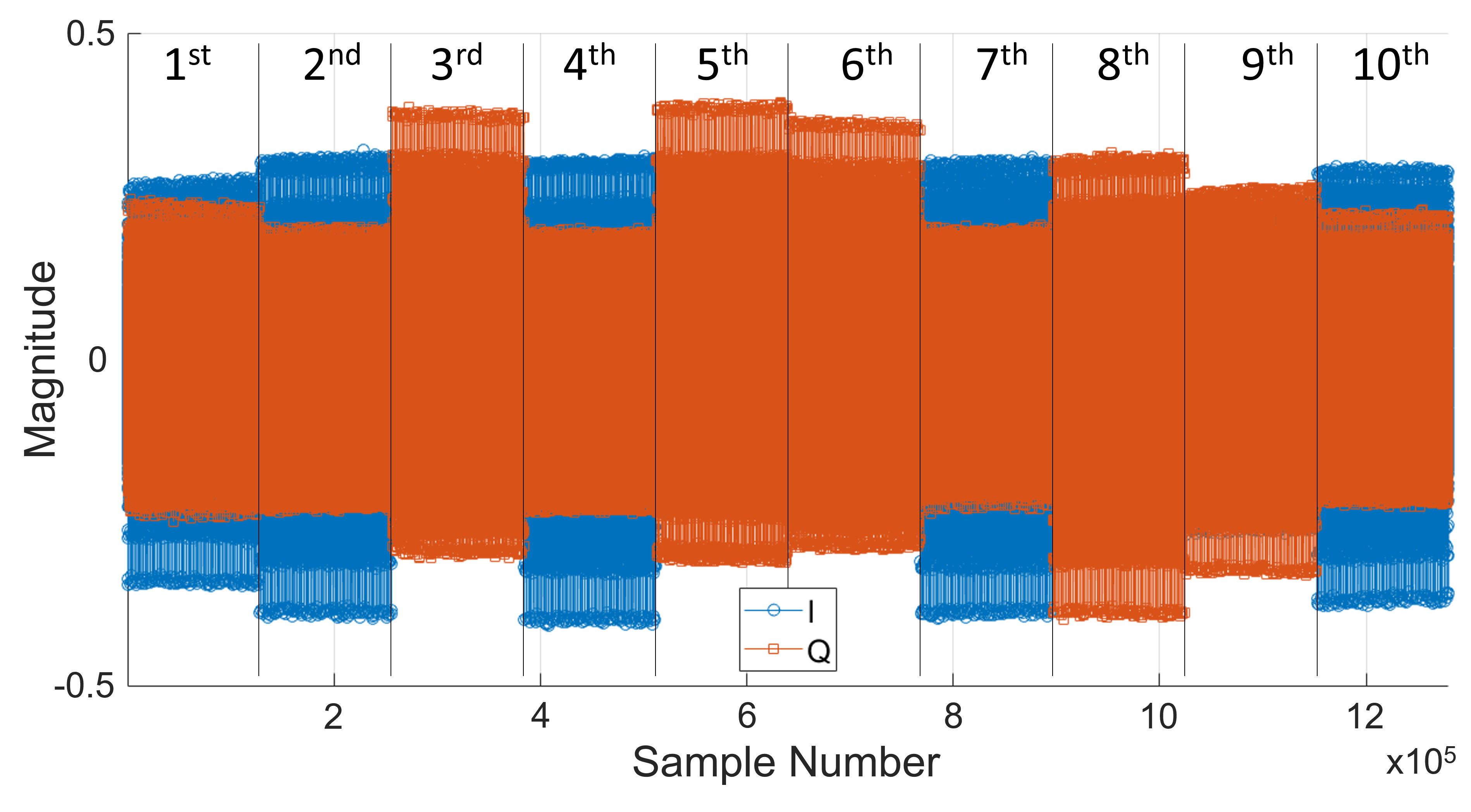}\vspace{-0.0cm}
  \caption{Ten concatenated acquisitions forming a multi-acquisition of the CC-free signal.}
  \label{fig:concat_data} \vspace{-0.0cm}
\end{figure}

As the received signal passes through different digital and analog hardware components, every received sample is affected by hardware impairments that impact the signal at baseband and RF, e.g., flicker noise, quantification error, DC offset (DCO), IQ imbalance (IQI), carrier frequency offset (CFO), phase noise, and jitter. Moreover, although the SMA connection reduces channel impairments, the test environment is not noise-free, thus the SMA cable loopback connection is affected by various noise sources, such as thermal noise or interference from other signals.
Such impairments in the dataset signals are not compensated for after acquisition. 
However, a single long acquisition of 2000 frames has similar hardware and RF impairments conditions for all received frames. A richer dataset should take into account diverse hardware and RF impairments conditions, as well as hardware temperature.
Hence, we performed ten smaller acquisitions of 200 frames and concatenated them to form a single 2000 frames multi-acquisition dataset element. In this way, each dataset element comprises signals with differently distributed non-idealities.
As an example, Fig. \ref{fig:concat_data} shows the concatenation of the 10 acquisitions of I/Q samples for the CC-free signal in the dataset.

\section{RF Transceiver Architecture with AI-based HT-CC detection}
\label{sec:Defense_CC}

AI-based HT-CC detection operates as a run-time, post-silicon defense. Because it detects abnormalities directly in raw received transmission frames caused by CC activity, it offers a generic solution capable of addressing the wide range of existing HT-CC attack models, their growing sophistication, and even unknown or undocumented variants. Rather than relying on multiple combinations of countermeasures, as outlined in Section \ref{sec:related_work}, AI-based HT-CC provides a unified final line of defense.

In \cite{D-RAAS24}, the emphasis is placed on designing the CNN model, but the work does not address where the model should reside and inference should be performed. We identify three possible locations for executing CNN inference: on the HT-infected RF transmitter (Alice), in the cloud which would require Alice or the nominal RF receiver (Bob) to forward the signal upstream, or on Bob itself, which would necessitate integrating a CNN hardware accelerator on-chip.

Placing the CNN model on Alice is not a secure option, since the same sophisticated attacker who inserted the HT could also tamper with the local CNN inference to suppress any alerts indicating CC activity.

On the other hand, having Alice (Bob) verifying the integrity of its transmitted (received) signal by retransmitting the signal to the cloud, where the CNN inference is performed, increases the power consumption of Alice (Bob). If the CC remains continuously active, only a small number of retransmissions are needed for detection, keeping the power overhead manageable. However, if the CC is enabled intermittently or for short durations, this approach would require near-continuous data retransmission, leading to impractically high power usage. Sending periodic frames to the cloud to check for CC activity could also allow the CC to operate undetected for long periods or even avoid detection entirely. Additionally, retransmission introduces privacy concerns and adds latency to CC detection.

Therefore, the most suitable location for executing the CNN model is within Bob’s RF receiver, which preserves data privacy and enables low-latency CC detection, assuming a lightweight CNN hardware accelerator is employed to balance performance and cost. 

This decision is supported by prior studies comparing cloud-based and edge-based AI execution in terms of energy consumption and latency. For instance, \cite{cost} reports that the energy cost of Large Language Model (LLM) inference is approximately 1.65 cents per response on a cloud platform, compared to just 0.0041 cents on an edge device. Similarly, \cite{latency2} presents large-scale latency measurements showing that median data transfer latency, based on Round-Trip Time (RTT), can reach up to 60 ms in cloud scenarios, whereas edge deployments reduce this to around 20 ms. The IoT seven-level reference model proposed in \cite{cloudlayer} depicts the multi-layered data flow from the edge device, through communication and edge computing layers, up to cloud-based processing and application layers, where each upward transition between each layer introduces additional latency and reduces privacy due to increased data transfers. This layered architecture between edge and cloud not only delays CC detection but also increases the risk of data leakage before any mitigation can occur. In contrast to cloud-based approaches, local detection on the edge device enables rapid, real-time detection and mitigation of CC threats.

\begin{figure}[t]
    \centering
    \includegraphics[width=1\linewidth,height=4cm]{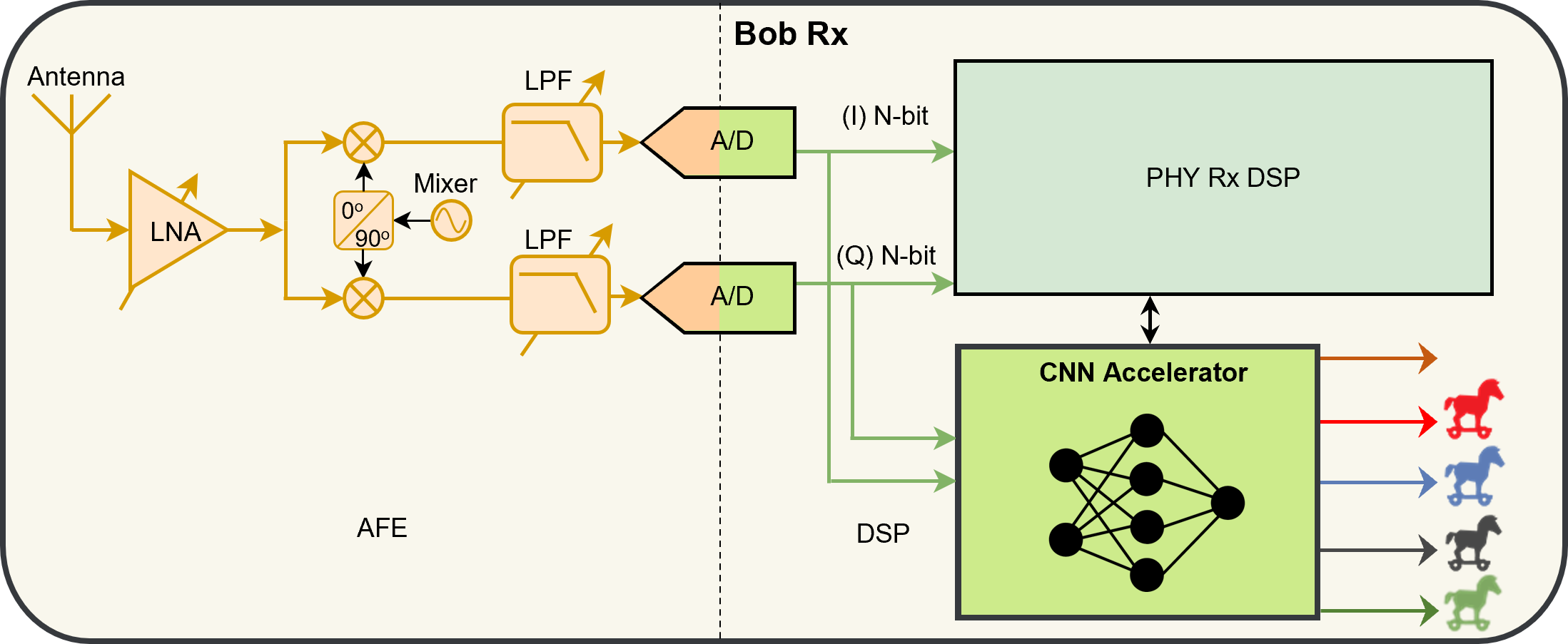}
    \caption{Bob's RF receiver architecture with AI-based HT-CC detection.}
    \label{fig:SoC_AI_PHY}
\end{figure}

Fig. \ref{fig:SoC_AI_PHY} shows the proposed localized CC detection approach. Bob’s RF receiver integrates a dedicated CNN accelerator, eliminating the need for external cloud processing and its associated latency. The accelerator continuously monitors the incoming digital I/Q samples directly at the ADC outputs of the AFE, operating in parallel with the RF receiver so that normal communication between Alice and Bob remains unaffected. By using the raw received signal as input, rather than relying on parametric features like power as in SCCF, the CNN performs direct CC analysis instead of side-channel analysis. Additionally, because the CNN observes the signal at the earliest possible digital stage, immediately after RF–to–I/Q conversion, it prevents sensitive information from propagating to higher-layer processing blocks such as the PHY or MAC. If a CC is detected, Bob immediately halts communication and notifies Alice, who may be unaware that sensitive data is being exposed to Eve.

\section{CNN Model}\label{sec:CNN}

\subsection{Baseline CNN model and classifier type exploration}\label{classifier_type_exploration}

As presented in Section \ref{sec:dataset}, the dataset contains for 8 different SNR values, ranging from 1dB to 29dB with a step of 4dB, $5 \times 2000$ frames labeled to 1 out of 5 classes, namely CC-free and HTX-CC, X=$\{1,\cdots,4\}$. The Wi-Fi physical layer frame contains $2 \times 640$ real-value samples corresponding to the two I/Q branches, as detailed in Section \ref{subsec:cover_signal}.

The classifier processes one frame at a time and produces a classification for each frame. We consider two classification problems: (a) Binary classification, which measures the ability to distinguish CC-infected transmissions from CC-free ones, with all HTX-CC variants, X=$\{1,\cdots,4\}$, grouped into a single CC-infected class; and (b) Multi-class classification, a more challenging task in which the classifier predicts 1 of 5 classes, CC-free or HTX-CC for X=$\{1,\cdots,4\}$. In this setting, the classifier not only detects whether a transmission is CC-infected but also identifies the specific HT mechanism used at the transmitter.

\begin{figure}[t]
    \centering
    \includegraphics[width=1\linewidth]{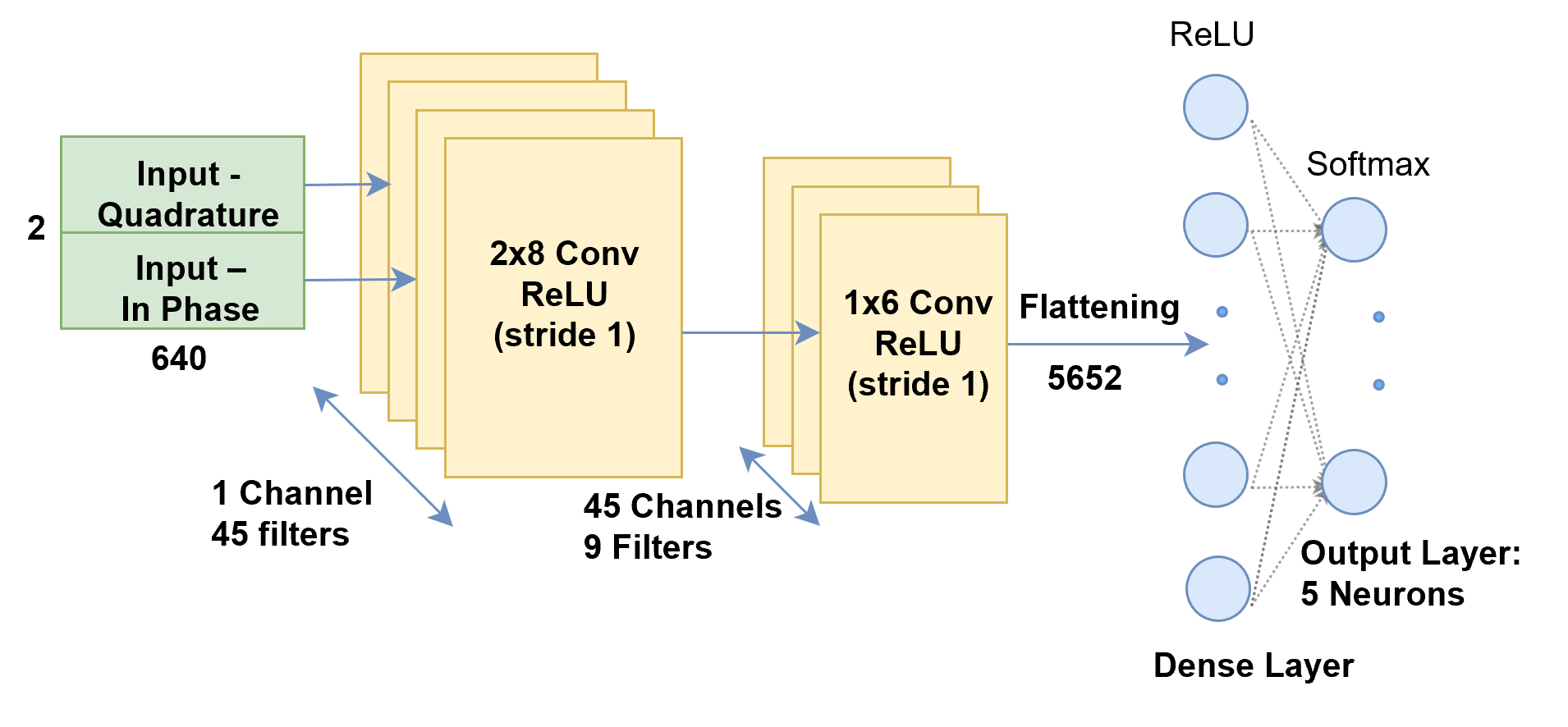}\vspace{-0.0cm}
    \caption{Baseline CNN model architecture.}
    \label{fig:baseline_cnn} \vspace{-0.0cm}
\end{figure}

In \cite{D-RAAS24}, the frame is encoded directly as a $2 \times 640$ ``image" and a CNN model is adopted for its proven effectiveness in image classification and robust feature-extraction capabilities. Starting from a large standard CNN architecture and following a trial-and-error approach, the number of layers and the layers' size were reduced while maintaining the maximum accuracy. The resultant baseline architecture, shown in Fig. \ref{fig:baseline_cnn}, consists of two convolutional layers, one fully connected layer, and an output layer, totaling 184,265 parameters ,e.g., synaptic weights. As discussed in detail in Section \ref{subsec:different_classifiers}, this baseline CNN model achieves average accuracies across SNR values of 90.9\% and 88.26\% for the binary and multi-class classification tasks, respectively.

\begin{figure*}[t]
    \centering
    \includegraphics[width=0.8\linewidth]{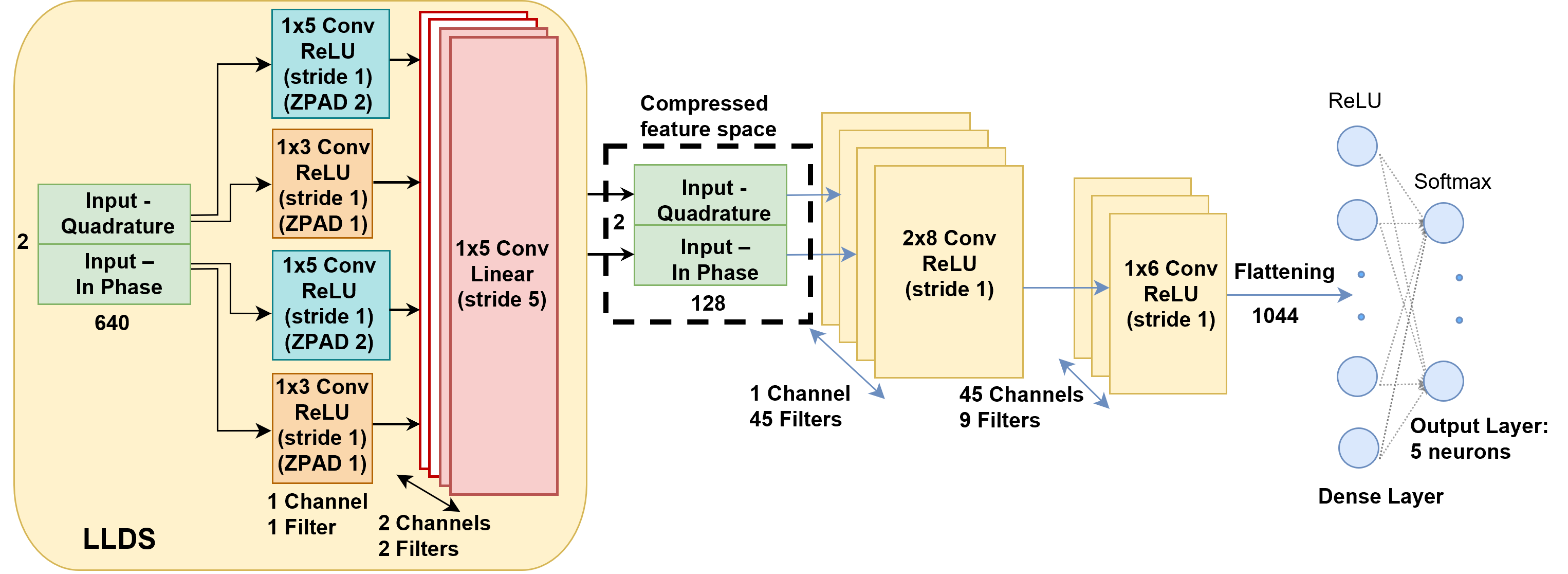}
    \caption{Proposed compact CNN model with feature-compression block at input.}
    \label{fig:cmodel}
\end{figure*}

Since our goal is to integrate the CNN hardware accelerator with the RF transceiver, our initial focus was to compact the CNN model without degrading the accuracy of the baseline model in \cite{D-RAAS24}, ensuring that the accelerator meets the area and power constraints of the wireless IC. As illustrated in Fig. \ref{fig:cmodel} and detailed in Section \ref{subsec:compressed_cnn}, this is accomplished by reducing the CNN model’s input dimensionality and introducing a feature-compression block at the CNN input, which maps the original $2 \times 640$ input into the reduced representation.

We also evaluate several alternative classifier types to motivate the choice of a CNN. As shown in Section \ref{subsec:different_classifiers}, their performance on the dataset confirms that the proposed compact CNN provides the most effective solution.

The alternative classifier types examined include the one-class SVM used in SSCF \cite{LJNM17,SHANM20}, a multi-class SVM, and two additional Deep Neural Networks (DNNs) beyond the CNN models, namely a Fully-Connected Neural Network (FCNN) and a Long Short-Term Memory (LSTM) network, the latter being a type of Recurrent Neural Network (RNN). The LSTM is considered because its strong ability to capture long-term dependencies makes it well suited for time-series data such as the continuously received I/Q frames.

The one-class SVM is an unsupervised method trained solely on CC-free frames (inliers) and subsequently used to detect CC-infected frames (outliers), making it inherently limited to binary classification. In contrast, the multi-class SVM, FCNN, and LSTM are employed as multi-class classifiers in the same manner as the CNN models.

The SVMs use the Radial Basis Function (RBF) kernel. The I/Q samples are concatenated to make a 1280-dimensional input, then PCA is applied keeping a number of principal components such that 90\% of data variation is explained. This results in a 253-dimensional input for the one-class SVM and a 320-dimensional input for the multi-class SVM. 

The FCNN and LSTM use a $2 \times 640$ input, similar to the CNN models. A trial-and-error approach was employed to design the most compact architectures that still preserve the maximum observed accuracy. The resulting FCNN comprises three dense layers with a total of 406,541 parameters, while the LSTM model consists of two LSTM layers followed by a dense layer and an output layer, totaling 451,200 parameters.

\subsection{Compact CNN model with feature-compression block}\label{subsec:compressed_cnn}

We opted for retaining the architecture of the baseline CNN while reducing its size by compressing the input dimensionality using a compression factor (CF) $CF=2 \times c$, $c \in \mathbb{Z}^+$, to $2 \times \frac{640}{CF}$. However, since the CC detection task operates on frames of dimension $2 \times 640$, an additional learnable block must be inserted between the frame input and the compressed CNN model to reduce the dimensionality from $2 \times 640$ to $2 \times \frac{640}{CF}$. This block and the compressed CNN are trained jointly in an end-to-end manner.


First, we experimented with max- and average-pooling layers to compress the frame length, but this approach led to a 
drop in accuracy with respect to the baseline model for any $CF$ value. Next, we considered using an autoencoder, which is commonly employed for feature compression. While it only caused a minor accuracy loss, 
the autoencoder contains approximately 164,000 parameters, conflicting with our original goal of minimizing model size for efficient on-chip implementation.

To address this challenge, we introduce a feature-compression block called Learnable Linear Down-Sample (LLDS), designed to reduce feature dimensionality, while incurring minimal overhead and preserving accuracy. The LLDS block precedes the CNN model, with the complete combined model shown in Fig. \ref{fig:cmodel} for $CF=5$, selected as the optimal $CF$ based on an ablation study discussed later. The LLDS block consists of four parallel single-convolution filters at its input (e.g., two per I/Q channel), followed by a down-sampling convolution layer. The parallel input convolutions capture multiple receptive fields, effectively enhancing informative samples while suppressing uninformative ones. The learned filters at this stage have sizes $1 \times 3$ and $1 \times 5$ with stride 1, drawing inspiration from the Inception architecture \cite{inception}. Zero padding (ZPAD) padding is symmetrically applied to left and right to preserve the input dimension of 640, resulting in an output of two channels with dimensions $2 \times 640$. The subsequent convolution layer applies two filters of size $1 \times 5$ across the two channels with stride equal to $CF$ (e.g., $CF=5$). This operation efficiently prunes redundant or non-informative features emphasized by the preceding layer. Consequently, the LLDS block outputs a reduced feature representation of dimension $2 \times \frac{640}{CF}$.

\begin{table}[t]
    \centering
    \scriptsize
    \caption{Ablation study for finding the optimal CF.}
    \label{tab:ablationstudy2}
    \begin{tabular}{p{1.5cm}p{1.2cm}p{1.2cm}p{1.2cm}p{1.2cm}}
        \toprule
           Compression Factor ($CF$) & Input dimensions of CNN model & Average binary-class accuracy drop   & Average multi-class accuracy drop & Parameters number  \\
          \midrule
          Baseline & 2x640           & 0.00\%  & 0.00\%& 184,265    \\
          CF = 2  & 2x320          & -0.25\%  & -0.30\%& 92,135    \\
          CF = 3  & 2x213          & -0.53\%  & -1.70\%& 61,323     \\
          CF = 4   & 2x153         & -0.55\%  & -1.70\%& 46,063  \\
          CF = 5      & 2x128      & -0.62\%  & -1.76\%& 36,851 \\
          CF = 6  & 2x106          & -0.70\%  & -2.50\%& 30,519  \\
        \bottomrule
    \end{tabular}\vspace{-0.0cm}
\end{table}

To identify the optimal $CF$ that balances classification accuracy and model size, we conducted an ablation study by varying $CF$, with the results summarized in Table~\ref{tab:ablationstudy2}. For each $CF$, the table reports the resulting CNN input dimension, the accuracy degradation relative to the baseline model for both binary and multi-class classification tasks, and the number of parameters in the CNN model which is compacted for $CF>1$. Here, we report average accuracies across SNR values. More detailed results will be presented in Section \ref{cnnresults}. Based on the findings in Table~\ref{tab:ablationstudy2}, we selected $CF=5$, which yields only a modest accuracy reduction of 0.62\% and 1.76\% for the binary and multi-class tasks, respectively, while reducing the input dimension to $2 \times 128$
and the number of parameters from 184,265 to 36,851, corresponding to an 80\% model compression.

  \begin{table}[t]
     \centering
     \caption{Ablation study of different activation functions in down-sampling layer in LLDS module.}
     \label{tab:ablation}
     \begin{tabular}{p{2cm}p{2.5cm}p{2.5cm}}
         \toprule
            Activation function & Average binary accuracy drop relative to linear activation & Average multi-class accuracy drop relative to linear activation\\
           \midrule
           Linear            &-0.00\%  &-0.00\% \\
           Tanh              &-1.24\%  &-3.35\% \\
           Sigmoid           &-1.87\%& -4.48\% \\
           ReLU              &-3.09\%  & -7.06\% \\
           Leaky-ReLU        &-3.59\%  & -7.10\% \\

         \bottomrule
     \end{tabular}\vspace{-0.0cm}
 \end{table}

We additionally performed an ablation study to assess the effect of different activation functions within the LLDS block.  Although feature-extraction CNN architectures typically use a single activation function throughout all layers, our focus was on varying the activation in the large second convolutional layer of the LLDS block. The results are summarized in Table \ref{tab:ablation} where we report the accuracy drop for the binary and multi-class classification problems relative to the linear case (i.e., no activation) which proved to be the best-performing choice. Surprisingly, using ReLU, i.e., the activation employed in the main CNN model, resulted in large accuracy degradation. We further evaluated Tanh, Sigmoid, and Leaky ReLU, finding that the linear case consistently delivered the best results. This suggests that preserving linearity during the down-sampling step retains more discriminative information, while nonlinear functions such as ReLU can suppress informative negative values and Sigmoid/Tanh may excessively constrain the feature space.

The LLDS block introduces negligible computational overhead, contributing only 3\% of the model’s total FLOPs and 0.11\% of its overall parameters. 


\section{CNN Hardware Accelerator Architecture}\label{acceleratorarchitecture}

\begin{figure*}[h!]
    \centering
    \includegraphics[width=0.8\linewidth]{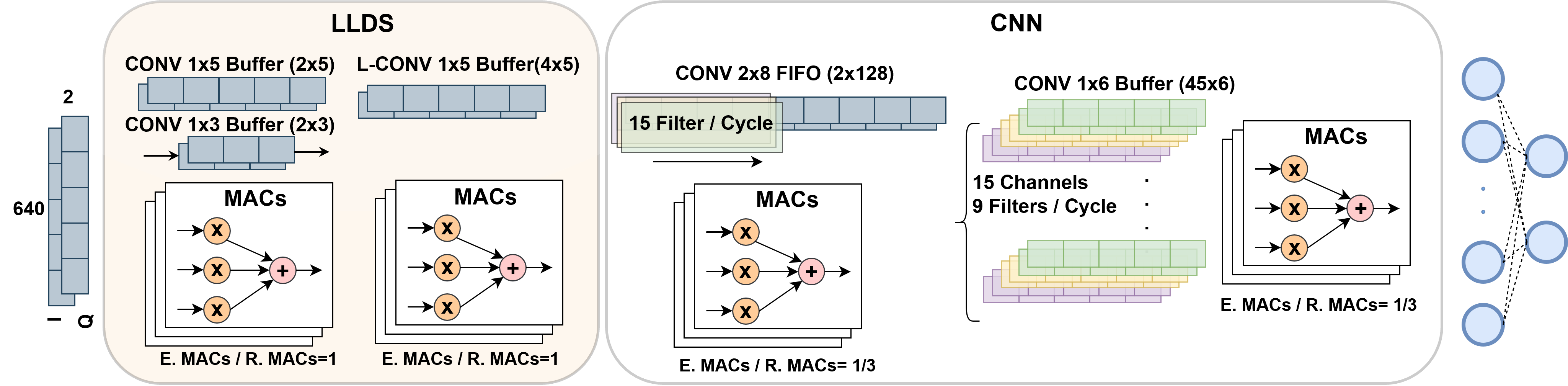}
    \caption{CNN accelerator data-flow architecture.}
    \label{fig:accelerator1}
\end{figure*}

\begin{figure}[t]
    \centering
    \includegraphics[width=0.8\linewidth]{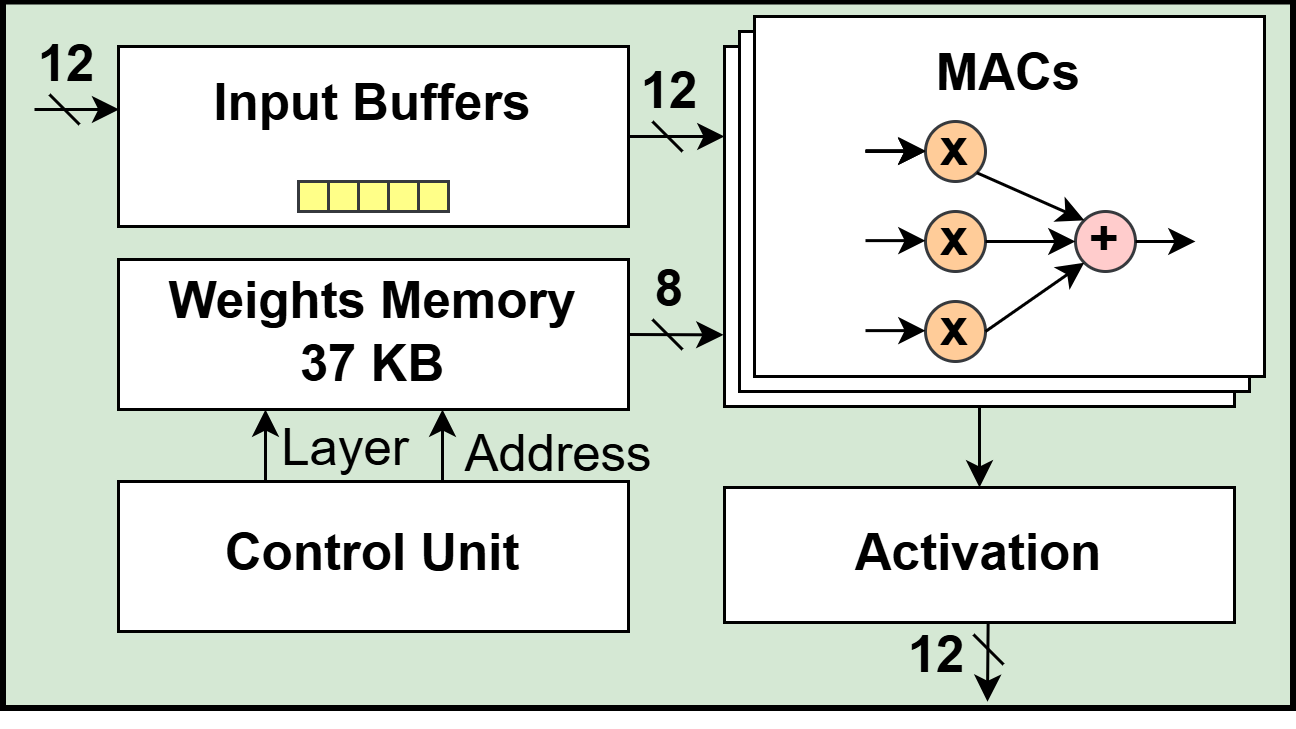}
    \caption{Layer architecture.}
    \label{fig:layerarch}
\end{figure}

The data flow of the proposed CNN hardware accelerator is illustrated in Fig.~\ref{fig:accelerator1}. The LLDS block is designed to keep filter weights stationary while sliding the input data at each clock cycle. This scheduling strategy enables the LLDS to operate at full speed alongside the RF receiver, remaining synchronized with the incoming frame rate. Such synchronization is critical to prevent frame loss, as the CC may be embedded in specific frames and dropping frames could result in the CC going undetected. Consequently, the LLDS block achieves maximum throughput by executing all required Multiply–Accumulate (MAC) operations in parallel, that is, the execution rate, defined as the ratio of executed MACs per clock cycle to the required MACs per clock cycle (E.~MACs/R.~MACs), is equal to 1. 

Additionally, this buffering strategy eliminates the need to buffer the large $2 \times 640$ input frame, significantly reducing Flip-Flop (FF) utilization on the FPGA by limiting intermediate pipeline stages and minimize amount of data transferred between layers. Overall, since the LLDS module introduces only a limited number of MAC operations, its contribution to FPGA resource utilization remains negligible.

In contrast, the CNN dominates the overall MAC count of the model. Therefore, its convolution filters are partitioned across the available multiplier and adder resources to preserve efficiency and minimize power consumption. Specifically, one third of the filters are processed per clock cycle, yielding an execution rate of E.~MACs/R.~MACs=1/3, as shown in Fig.~\ref{fig:accelerator1}, and resulting in an overall processing latency of three input frame durations. For example, in the first convolution layer, which comprises 45 filters, 15 filters are executed per clock cycle, completing the filter shifts per channel in three clock cycles.

With the LLDS module compressing the input dimensionality by $CF=5$, 
buffering requirements at the input of the CNN are significantly relaxed, allowing the LLDS output to be stored efficiently using a FIFO, in order to accommodate the rate transition between the execution rate of the LLDS module (E.~MACs/R.~MACs=1) and the execution rate of hte CNN (E.~MACs/R.~MACs=1/3). With each convolution shift being processed over three clock cycles, the incoming compressed frame from the LLDS block is stored entirely in the FIFO to avoid any loss of adjacent samples during the execution of each convolution shift. As a consequence of that, unlike the LLDS stage, convolution layers of the CNN employ an input-stationary data-flow, in which input feature maps are held stationary while filter weights slide over them, as illustrated in Fig.~\ref{fig:accelerator1}.

Fig.~\ref{fig:layerarch} illustrates the internal layer architecture. Since the dataset consists of real-valued I/Q samples, a 12-bit quantization is applied to reflect the typical resolution of ADCs used in RF receiver front-ends. In addition, post-training quantization of the filter weights to 8 bits is performed. As shown in Section \ref{cnnresults}, these quantizations introduce only a minor accuracy degradation compared to the software model. The control unit manages weight-memory addressing and aligns the input data with the corresponding weights, coordinating with dedicated MAC units. These MAC units execute fixed-point 12-bit $\times$ 8-bit operations, after which the results are passed through the activation function and truncated back to 12 bits.

\section{Results}\label{results}

\subsection{SNR requirements for HT-CC error-free recovery and HT-CC detection}
\label{SNR_requirements}

\begin{table*}
    \centering
    \caption{Minimum SNR for HT-CC error-free recovery and for HT-CC detection with accuracy $>90\%$ using the CNN model.}
    \label{tab:min_SNR_CC_model}
    \begin{tabular}{p{1.8cm}p{8cm}p{7cm}}
       \toprule
       Attack model & Minimum SNR required for error-free recovery of the covert data & Minimum SNR required for HT-CC detection using the CNN with an accuracy exceeding 90\%\\
       \midrule
       HT1-CC \cite{DiRiAbSt22} & 24~dB (without any error correction techniques) & 11~dB\\
       HT2-CC \cite{ClScHo15} & 35~dB (for 256-PSK having 0.1\% cover channel BER) & 1~dB\\
       HT3-CC \cite{Dutta13} & 24~dB (for 98\% packet reception) & 14~dB\\
       HT4-CC \cite{SHANM20} & 30~dB (for a BPSK modulation and a CC rate of 5 bps) & 15.5~dB\\
       \bottomrule
    \end{tabular}\vspace{-0.0cm}
\end{table*}

Before presenting the results, it is important to distinguish the SNR range in which an attacker can achieve error-free recovery of HT-CC data, as this also defines the SNR range in which HT-CC detection is truly necessary. If the attacker cannot reliably decode the leaked information, then HT-CC detection performance in that SNR regime is of limited practical relevance.

The dataset spans SNR values from 1~dB to 29~dB. However, robust and reliable Wi-Fi communication typically requires a minimum SNR of 20–25~dB \cite{ieee80211}. Table \ref{tab:min_SNR_CC_model} shows, in its second column, the minimum SNR needed for error-free covert data recovery (without redundancy) for the four HT-CC schemes of the dataset listed in the first column, as reported in their respective publications. These thresholds indicate that an attacker cannot decode the leaked information without errors at low SNR levels. Consequently, HT-CC detection must be highly reliable in high SNR conditions, whereas optimizing detection at low SNR values is less critical, as even the cover communication suffers from high BER in that regime.

Although the relevant SNR range for the attacker, where leaked data can be correctly recovered and HT-CC detection is therefore meaningful, lies above 20~dB, we still report results down to 1~dB. This allows us to assess the robustness of HT-CC detection under challenging, low-SNR conditions where communication quality is inherently poor.


\subsection{Results of baseline CNN model and classifier exploration}
\label{subsec:different_classifiers}

\begin{figure}[t]
	\centering
	\includegraphics[width=0.9\linewidth]{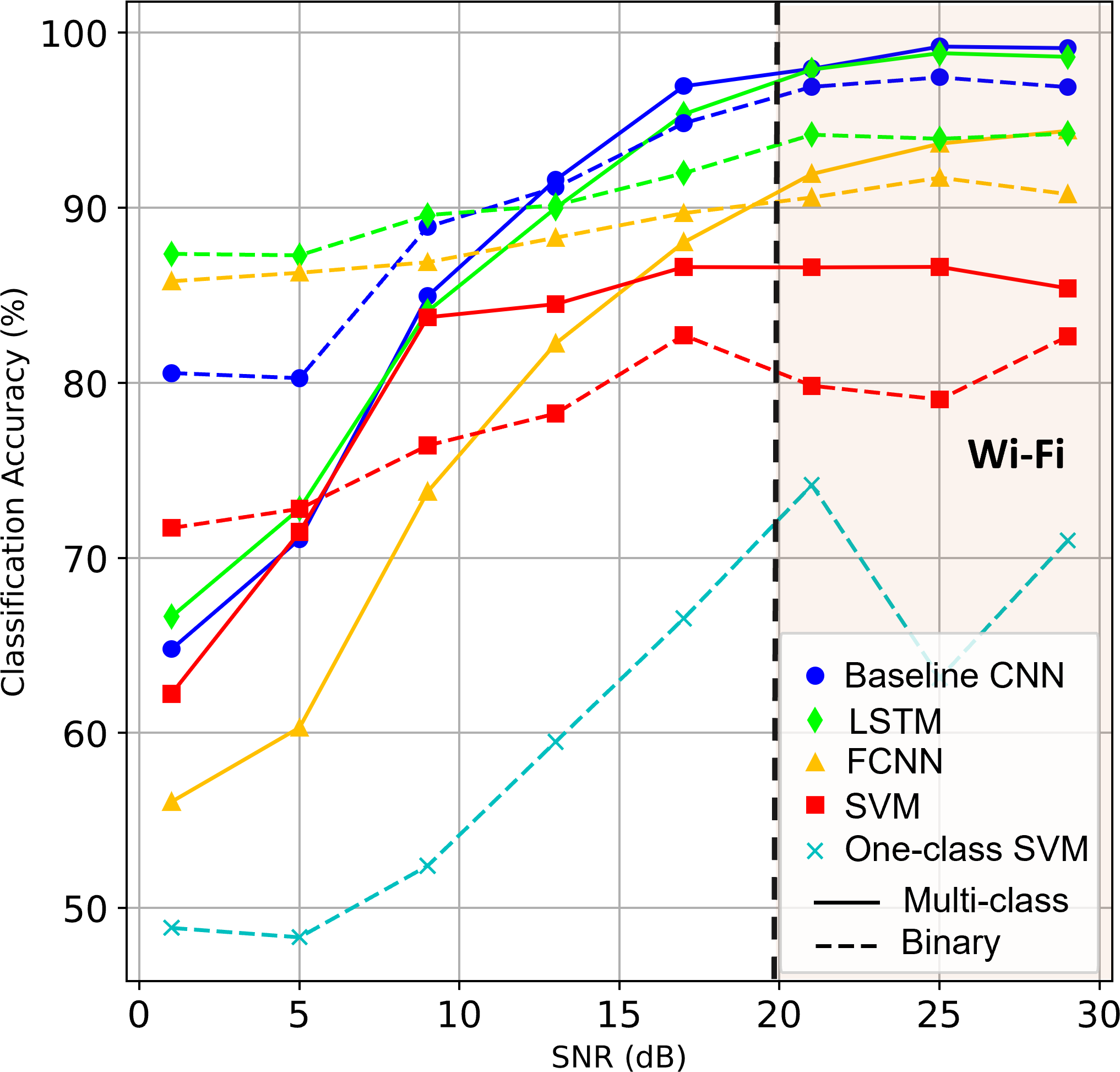}
	\caption{Classifier performance comparison as a function of SNR.}
	\label{figure:svmcnn} \vspace{-0.0cm}
\end{figure}

Fig. \ref{figure:svmcnn} shows the performance comparison of the 5 classifiers presented in Section \ref{classifier_type_exploration}, namely one-class SVM \cite{LJNM17,SHANM20}, multi-class SVM (which is denoted as simply SVM), baseline CNN \cite{D-RAAS24}, FCNN, and LSTM, evaluated on the dataset across varying SNR levels. 

 \begin{table}[t]
    \centering
    \scriptsize
    \caption{Input dimensionality, number of parameters, and average classification accuracy on the dataset over the SNR range for the different classifiers.}
    \label{tab:models}
    \begin{tabular}{p{1.6cm}p{0.8cm}p{1.2cm}p{1.1cm}p{1.1cm}}
        \toprule
           Model & Input size  & Parameters number & Average binary accuracy & Average multi-class accuracy \\
          \midrule
          Baseline CNN              & 2x640 & 184,265   &  90.9\%& 88.26\%\\
          LSTM             & 2x640 & 451,200   & 91.10\% & 88.09\% \\
          FCNN             & 1280  & 406,541   &  88.70\% & 80.10\%\\
          SVM              & 320   & -         &  77.40\% & 80.93\%\\
          One-class SVM    & 253   & -         & 66.40\% & -\\
          LLDS+CNN    & 2x640   & 36,851         & 90.28\% & 86.50\%\\
          LLDS+CNN (Quantized)   & 2x640   & 36,851         &90.27\%  & 85.80\%\\
        \bottomrule
    \end{tabular}\vspace{-0.0cm}
\end{table}

Table \ref{tab:models} summarizes, for each classifier, the input dimensionality and number of parameters in the second and third columns, respectively (see Section \ref{classifier_type_exploration}). The fourth and fifth columns report the average accuracy on the dataset for the binary and multi-class classification tasks, respectively, averaged over all SNR values.

As shown in Fig. \ref{figure:svmcnn}, the trend is that the classification accuracy for both learning problems increases with SNR. This monotonic behavior is explained by the observation that the stronger the signal is, the easier it becomes for the classifier to spot the CC within the signal.

The one-class SVM shows the poorest accuracy on the binary classification problem that never exceeds 75\% and on average across all SNR values is 66.4\%. FCNN outperforms the SVM in the range of SNR$>$20~dB, which is the range of interest (see Section \ref{SNR_requirements}). This proves that a deep learning approach is rather needed for detecting a CC within raw transmission data. CNN and LSTM show the best accuracy outperforming all the other classifiers for both binary and multi-class classification problems. In the SNR range of interest, they show similar excellent accuracy for both classification problems that is over 97\% for SNR$>$20~dB and rises above 99\% for SNR$>$25~dB which is required for practical communication. 
Even for the lowest SNR of 1~dB, where the CC data cannot be recovered reliably by the attacker, they still perform surprisingly well achieving an accuracy for the binary classification problem of over 81\%, with the LSTM performing better than the CNN. For the multi-class classification problem, the curves of the LSTM and CNN are very close throughout the SNR range, while, starting from SNR=13~dB, the curves for both classification problems practically converge to an accuracy of over 85\%. For low SNR, the accuracy for the multi-class classification problem drops to around 65\% for SNR=1~dB and is over 85\% for SNR=10~dB. But this is a secondary issue since the primary goal is to detect the CC, while detecting the underlying HT mechanism within the transmitter is an auxiliary benefit. Furthermore, we observe that the CNN performs better in the range of SNR$>$20~dB and on average across the complete SNR range it achieves binary and multi-class accuracies very close to those of the LSTM. Therefore, given that the range of interest is SNR$>$20~dB, and given that the CNN has more than half less parameters than the LSTM as shown from Table \ref{tab:models} leading to a more efficient edge computing implementation, we conclude that the CNN is arguably the best choice towards an AI-based defense against HT-CCs.

\subsection{Results for compact CNN model with feature-compression block}
\label{cnnresults}

Herein, we report the performance of the compact CNN model, evaluated in both software and on the CNN hardware accelerator, and compare it against the baseline model.

 \begin{figure}[t]
   \centering
   \includegraphics[width=1\linewidth]{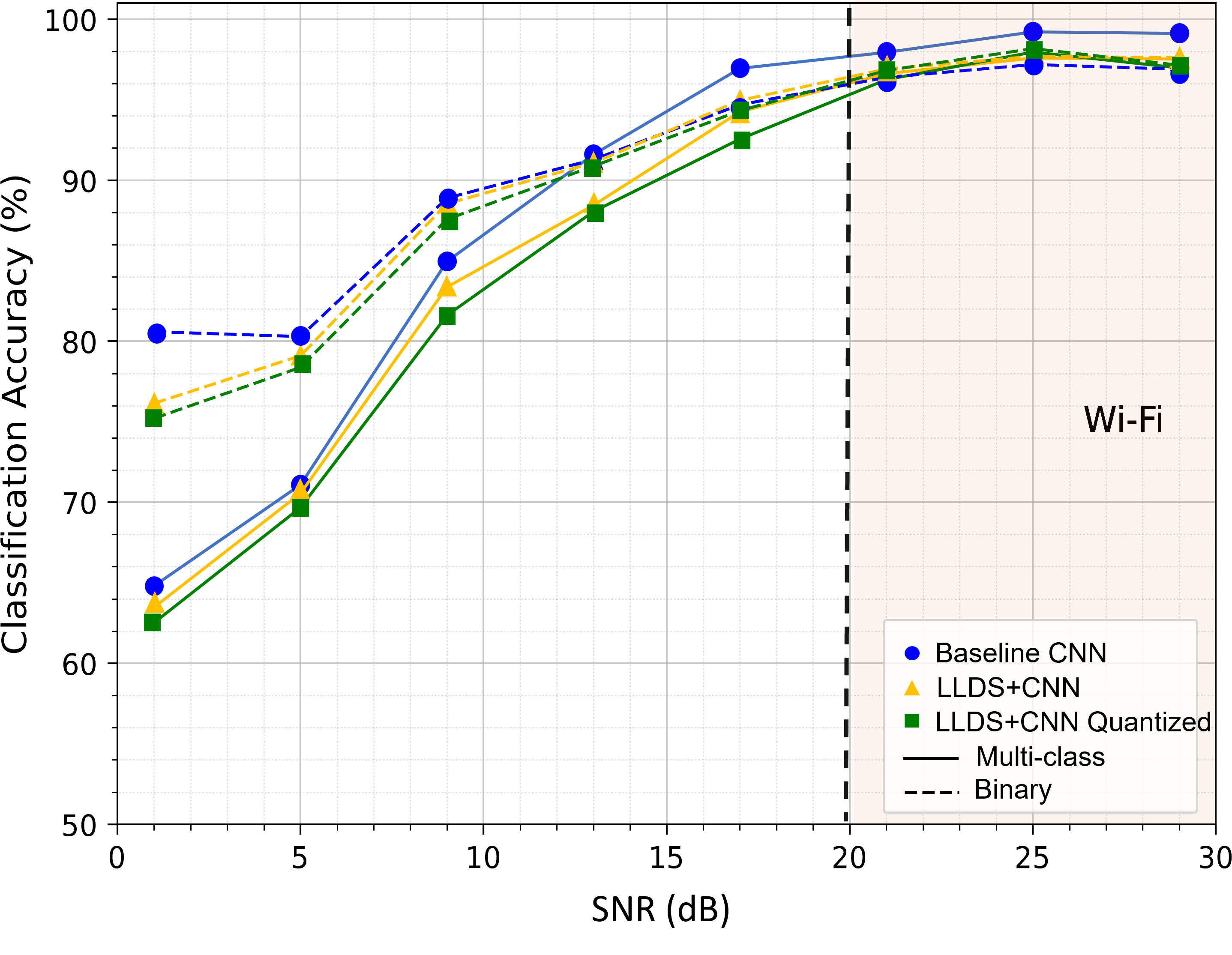}
   \caption{Baseline and compact CNN model performance comparison as a function of SNR.}
   \label{fig:fpgaacc}
 \end{figure}

Fig. \ref{fig:fpgaacc} presents the comparison across different SNR values, while the results of the compact model are added in the last two rows of Table \ref{tab:models}, where ``quantized" refers to the compact model with 8-bit weight precision executed on the hardware accelerator. 

In software, the compact model achieves average accuracies of 90.28\% and 86.5\% for the binary and multi-class classification tasks, respectively, compared to 90.9\% and 88.26\% for the baseline model, representing a small drop of less than 2\% for multi-class classification and less than 1\% for binary classification. This reduction comes with a substantial decrease in model size, from 184,265 to 36,851, enabled by the LLDS module parameters, corresponding to a 80\% reduction, which is crucial for efficient edge deployment. The 8-bit weight quantized compact model with 12-bit quantized I/Q samples running on the FPGA performs closely to its software counterpart for the binary classification task, and shows an average accuracy loss of less than 2\% for the multi-class classification task, which disappears in the SNR range of interest above 20~dB, as shown in Fig. \ref{fig:fpgaacc}.

\begin{figure}[t]
	\centering
	\includegraphics[width=1\linewidth]{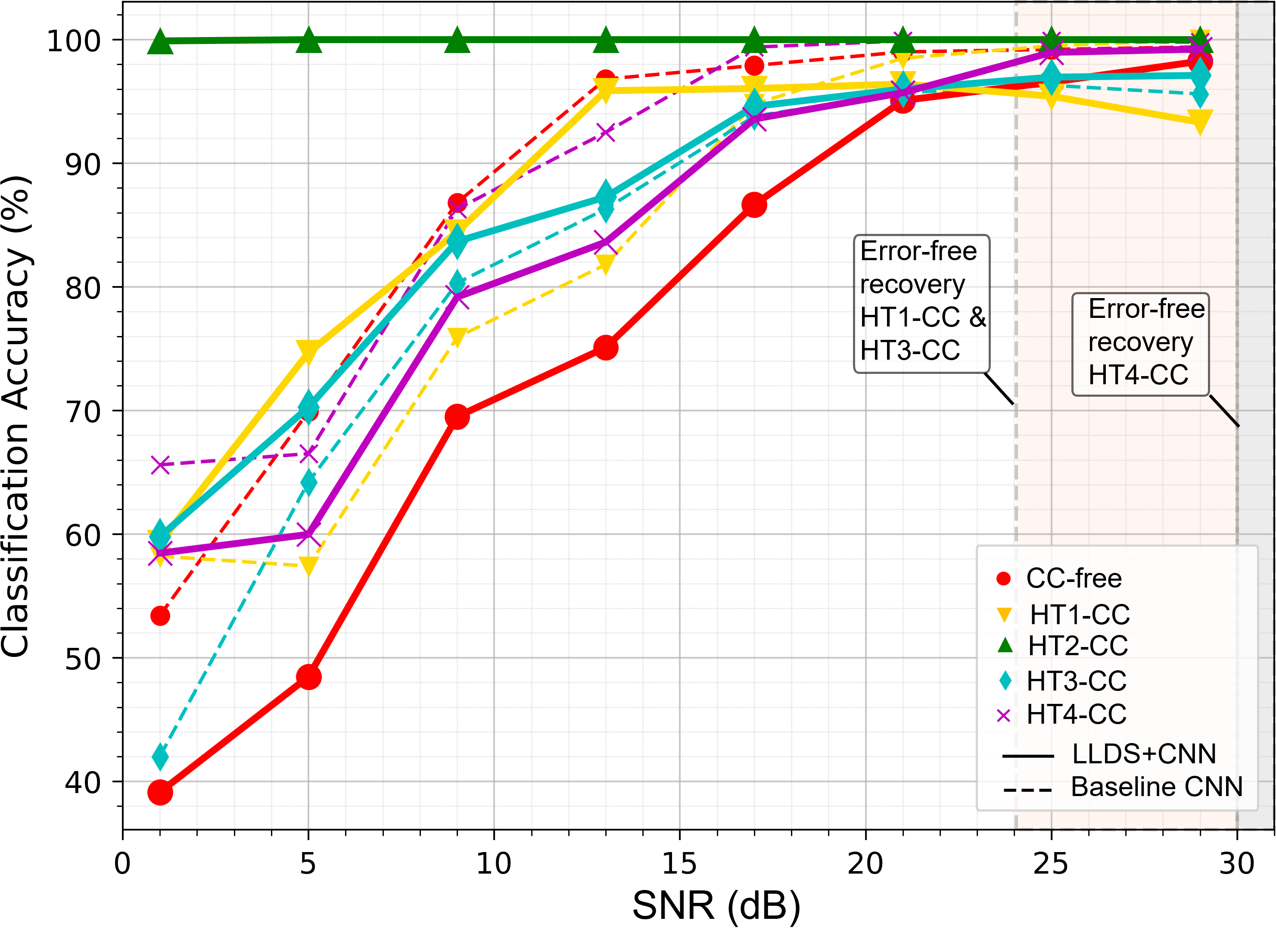}
    \vspace{-0.2cm}
	\caption{Baseline and compact CNN model performance comparison per class as a function of SNR.}
	\label{figure:class accuracy} \vspace{-0.0cm}
\end{figure}

Fig. \ref{figure:class accuracy} presents the per-class accuracies for the multi-class classification task. In the SNR range of interest above 20~dB, where an attacker can reliably recover the CC data, the compact model preserves the baseline model’s performance, achieving over 92\% accuracy for all classes, and over 96\% accuracy for the HT2-CC, HT3-CC, and HT4-CC classes. Notably, the accuracy remains high even at lower SNR values, highlighting the robustness of the AI-based detection method under challenging channel conditions. The last column of Table \ref{tab:min_SNR_CC_model} reports the minimum SNR required to achieve more than 90\% accuracy, as determined from Fig. \ref{figure:class accuracy}. As shown, accuracy stays above 90\% down to SNR values at least 10~dB lower that the minimum SNR required by the attacker for error-free CC recovery. At low SNR values, the compact model surpasses the baseline in detecting the HT1-CC and HT3-CC classes, while both models perform similarly for HT2-CC. The largest discrepancy appears in the CC-free class, where the compact model exhibits noticeably lower accuracy. In this case, CC-free samples are misclassified as HTX-CC, resulting in false alarms, an outcome that is arguably less critical than failing to detect an actual CC.

\begin{figure}[t]
    \centering
    \includegraphics[width=0.85\linewidth]{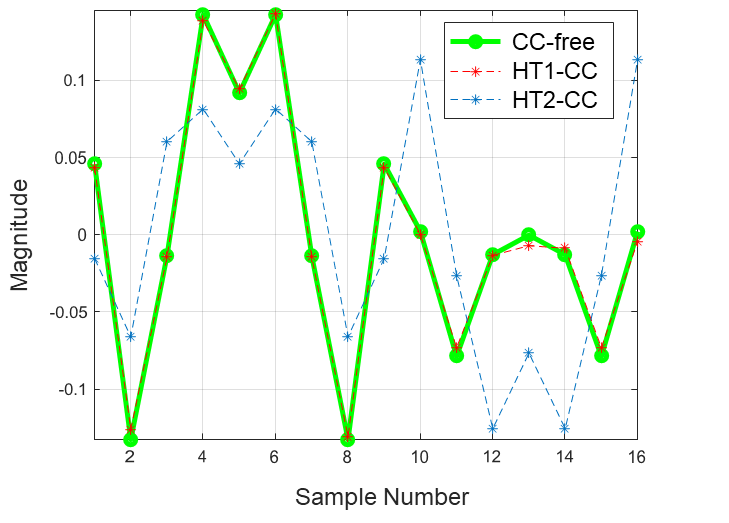}
    \caption{First 16 real (I) samples of the STS\textsubscript{t} for the CC-free and for HT1-CC and HT2-CC leaking a random byte.}
    \label{fig:sts_t}\vspace{-0.0cm}
\end{figure}

One notable observation from Fig. \ref{figure:class accuracy} is that HT2-CC is significantly easier to detect than the other classes. The detection accuracy remains at 100\% across the entire SNR range, even at the lowest SNR of 1~dB. This distinct behavior stems from the fact that HT2-CC produces a much more pronounced alteration in the ``image” of the transmitted frame compared to the other HT-CC variants, which makes it easily detectable by a CNN model. Specifically, HT2-CC applies a phase shift to all STS\textsubscript{F} symbols, as illustrated in Fig. \ref{fig:HT1_HT2}. After undergoing the IFFT, the phase-shifted STS\textsubscript{F} transforms into the time domain, modifying both the amplitude and phase of the resulting STS\textsubscript{t}. Fig. \ref{fig:sts_t} shows the first 16 I samples of STS\textsubscript{t} for the CC-free and the HT1-CC and HT2-CC cases when leaking a random byte, where HT2-CC stands out clearly as easily distinguishable.

\begin{figure}[t]
		\centering
		\includegraphics[width=0.8\linewidth]{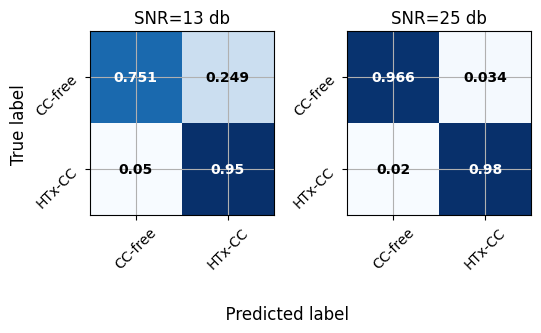}\vspace{-0.0cm}
		\caption{Confusion matrix of the compact CNN model for the binary classification task.}
		\label{figure:confusion_binaryLLDS} \vspace{-0.0cm}
	\end{figure}

\begin{figure}[t]
		\centering
		\includegraphics[width=1\linewidth]{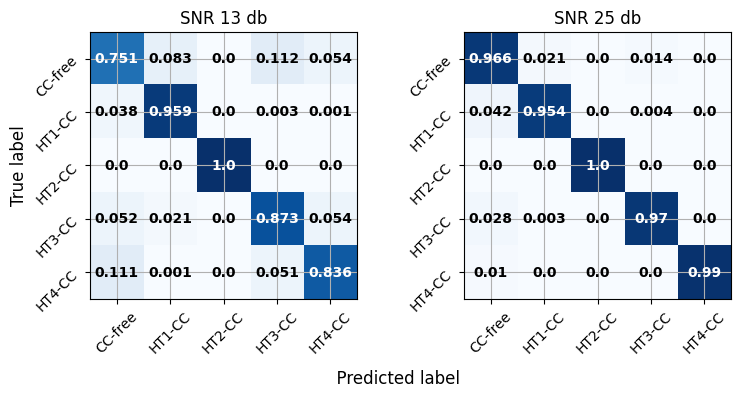}
		\caption{Confusion matrix of the compact CNN model for the multi-class classification task.}
		\label{figure:confusion_all_classesLLDS} \vspace{-0.0cm}
	\end{figure}

To shed light into the misclassification behavior, Figs. \ref{figure:confusion_binaryLLDS} and \ref{figure:confusion_all_classesLLDS} present the confusion matrices of the compact CNN model for the binary and multi-class classification tasks, respectively. To illustrate performance under varying channel conditions, we include results at two representative SNR levels. The first, 13~dB, is below the reliable Wi-Fi operating range and thus exposes inter-class ambiguity and misclassification patterns. The second, 25~dB, falls within the reliable reception range, where the classifier is expected to produce stable predictions. As seen, the confusion matrices at 25~dB are strongly diagonal, confirming this behavior. Notably, the diagonal structure is largely preserved even at 13 dB. At this lower SNR, the CC-free class exhibits the poorest detection accuracy, as also discussed above, being misclassified as HT-CC1 (8.3\%), HT-CC3 (11.2\%), and HT-CC4 (5.4\%).

\section{Comparison with other accelerators}
\label{compare_with_others}

\begin{table*}[t]
    \begin{center}
    \setlength\extrarowheight{3pt}
    \caption{Comparison of state-of-the-art FPGA implementations of AI hardware accelerators for RF signal classification tasks.}
     \setlength{\tabcolsep}{1pt}
    \begin{tabular}{>{\raggedright\arraybackslash}p{1.5cm} c c c c c c c c }
\hline
  & 
  \cite{Emad21} &
  \cite{soltani} &
  \cite{ISCAS25} &
  \cite{IEEEAccessSTFT} &
  \cite{Zhang} &
  \cite{Guo} &
  \cite{Gan}  &
  \textbf{This Work} \\
\hline
\textbf{Application}  &
Modulation &
Modulation & 
Modulation & 
Modulation & 
Modulation & 
Modulation &
Specific-Emitter &
\textbf{Covert} \\
 & Recognition &
 Recognition &
 Recognition&
 Recognition &
 Recognition &
 Recognition & 
 identification &
 \textbf{Channel} \\
\hline
\textbf{FPGA}  & 
XCZU7EV &
XCZU9EG&
XCZU7EV &
XCZU9EG&
XCZU5EG &
PYNQ &
XC7Z045 &
\textbf{XCZU5EG} \\
\hline
\textbf{Clock\newline(MHz)}  &
70 &
- &
115 &
{250} &
200 &
137 &
- &
\textbf{200} \\
\hline
\textbf{NN}  &
CNN &
ANN &
Attention + CNN & 
STFT-CNN &
CNN &
SNN &
Conditional &\textbf{LLDS}\\
&  & & &  &  &  & reconfigurable CVNN & \textbf{+ CNN}\\
\hline
\textbf{Weights\newline(\# of Bits)}  &
16 &
16 &
8 &
16 &
8 &
16 &
- &\textbf{8}\\
\hline
\textbf{LUT}  &
74.68K & 
158.43K &
75.36K &
97.90K &
67.77K &
31.73K &
- &
\textbf{27.07K} \\
\hline
\textbf{FF}  &
57726 &
16222 &
88623 &
139200 &
- &
50934 &
- &
\textbf{8246} \\
\hline
\textbf{DSP} &
1116 &
210 &
1728 &
578 &
131 &
0 & 
&\textbf{1229}\\
\hline
\textbf{Power\newline(mW)} &
847 &
1152 &
1192 &
10500 &
{858} &
2167 &
2150 &
\textbf{810}  \\
\hline
\textbf{Performance\newline(GOPs)}  &
33 &
15 &
{61} &
179 &
23 &
79 &
189 &
\textbf{86.5}\\
\hline
\textbf{Efficiency\newline(GOPs/W)} &
39 &
13 &
51 &
17 &
27 &
35 &
88 &
\textbf{107} \\
\hline
\end{tabular}
\\
    \label{table:hardware_results}
      \end{center}
\end{table*}

As this is the first AI hardware accelerator demonstrated for on-chip AI-based CC detection, Table \ref{table:hardware_results} compares our work with previously reported state-of-the-art FPGA-based AI hardware accelerators that explore a range of neural network architectures for RF signal classification tasks, particularly modulation recognition and emitter identification. For each design, we report the weight quantization, FPGA resources, power consumption, performance measured as the throughput, and energy efficiency expressed in Giga Operations/s (GOPs)

Operating at 200~MHz, the proposed accelerator delivers a throughput of 67~Mega Samples/s, corresponding to 43~Giga MAC/s or 86.5~GOPs (since each MAC consists of two operations), while consuming only 810~mW. This translates to an energy efficiency of 107~GOPs/W, ranking the accelerator as the most efficient design in Table~\ref{table:hardware_results}.

With respect to FPGA resource utilization, the proposed accelerator significantly reduces FF usage, requiring only 8,246 FFs, which is enabled by the weight-stationary data-flow of the LLDS block and the associated feature dimensionality reduction it provides.

\section{Conclusion}
\label{sec:conclusion}

We proposed a lightweight 810~mW 107~GOPs/W CNN hardware accelerator for on-chip real-time CC detection. The accelerator processes raw I/Q samples directly at the RF receiver ADC output, encoded as images, and operates at speed with the RF receiver. To enable efficient edge deployment, we first compressed the state-of-the-art CNN model in \cite{D-RAAS24} by 80\% with only a minor accuracy degradation, still achieving over 96\% CC detection accuracy and correct HT type classification within the SNR range of interest. We further justified the choice of a CNN for this task by benchmarking it against alternative classifier architectures, demonstrating a superior trade-off between model size and accuracy. All classifiers were trained using the hardware-generated HT-CC dataset from \cite{D-RAAS24}, which encompasses all major CC types, thereby demonstrating that the proposed strategy enables a unified, CC type-independent AI-based detection solution, in contrast to approaches requiring multiple dedicated countermeasures tailored to individual CC types.

\bibliographystyle{IEEE}
\bibliography{refs.bib}

\end{document}